\pdfoutput=1
\def\vdaarxiv{}
\def\vdaarxivheader{Preprint}
\def\vdaarxivrheader{Video Diffusion Avatars $\cdot$ Giebenhain et al.}

\documentclass{article}
\newif\ificlr\iclrtrue
\PassOptionsToPackage{dvipsnames}{xcolor}
\usepackage{iclr2027_conference,times}
\ifdefined\vdaarxiv\iclrfinalcopy\fi
\newcommand{\vdaprojecturl}{https://simongiebenhain.github.io/video-diffusion-avatars/}
\newcommand{\vdacodeurl}{https://github.com/SimonGiebenhain/video-diffusion-avatars}
\newcommand{\vdasuppvideo}{supplementary video\ifdefined\vdaarxiv{} on our \href{\vdaprojecturl}{project page}\fi}
\usepackage[T1]{fontenc} %

\usepackage{xcolor}
\usepackage{graphicx}
\usepackage{amsmath,amssymb}
\usepackage{booktabs}
\usepackage{xspace}
\usepackage[format=plain,labelformat=simple,labelsep=colon,font=small]{caption}
\usepackage[font=footnotesize,skip=3pt,subrefformat=parens]{subcaption}
\usepackage[hyphens]{url}
\usepackage[shortlabels,inline]{enumitem}
\setlist[itemize]{leftmargin=*,itemsep=1pt,topsep=2pt}

\input{preamble}

\usepackage{etoolbox,adjustbox}
\makeatletter
\newcommand{\iclr@fittabulars}{%
  \let\iclr@origtabular\tabular
  \let\endiclr@origtabular\endtabular
  \renewenvironment{tabular}{\begin{adjustbox}{max width=\linewidth}\begin{iclr@origtabular}}{\end{iclr@origtabular}\end{adjustbox}}}
\AtBeginEnvironment{table*}{\iclr@fittabulars}

\newcommand{\iclr@nocaptions}{\iclrnofloats\renewcommand\caption[2][]{}%
  \let\label\@gobble\def\vspace{\@ifstar\@gobble\@gobble}}
\newsavebox\iclr@boxA \newsavebox\iclr@boxB
\newlength\iclr@H \newlength\iclr@wA \newlength\iclr@wB \newlength\iclr@gap
\newcommand{\iclr@measure}[2]{\sbox#1{\iclr@nocaptions\input{#2}}}
\newcommand{\iclr@scaletabulars}{%
  \let\iclr@origtabular\tabular \let\endiclr@origtabular\endtabular
  \renewenvironment{tabular}{\begin{adjustbox}{totalheight=\iclr@H}\begin{iclr@origtabular}}{\end{iclr@origtabular}\end{adjustbox}}}
\newcommand{\iclrtablepair}[2]{%
  \iclr@measure\iclr@boxA{#1}\iclr@measure\iclr@boxB{#2}%
  \setlength\iclr@gap{1.2em}%
  \setlength\iclr@H{\fpeval{(\linewidth-\iclr@gap)/(\wd\iclr@boxA/(\ht\iclr@boxA+\dp\iclr@boxA)+\wd\iclr@boxB/(\ht\iclr@boxB+\dp\iclr@boxB))}pt}%
  \setlength\iclr@wA{\fpeval{\wd\iclr@boxA*\iclr@H/(\ht\iclr@boxA+\dp\iclr@boxA)}pt}%
  \setlength\iclr@wB{\fpeval{\wd\iclr@boxB*\iclr@H/(\ht\iclr@boxB+\dp\iclr@boxB)}pt}%
  \noindent\begin{minipage}[b]{\iclr@wA}{\iclrnofloats\iclr@scaletabulars\captionsetup{justification=raggedright,singlelinecheck=false}\input{#1}}\end{minipage}\hfill
  \begin{minipage}[b]{\iclr@wB}{\iclrnofloats\iclr@scaletabulars\captionsetup{justification=raggedright,singlelinecheck=false}\input{#2}}\end{minipage}}
\newcommand{\iclrfigtablepair}[2]{%
  \iclr@measure\iclr@boxA{#1}\iclr@measure\iclr@boxB{#2}%
  \setlength\iclr@gap{1.2em}%
  \setlength\iclr@H{\fpeval{(\linewidth-\iclr@gap)/(\wd\iclr@boxA/(\ht\iclr@boxA+\dp\iclr@boxA)+\wd\iclr@boxB/(\ht\iclr@boxB+\dp\iclr@boxB))}pt}%
  \setlength\iclr@wA{\fpeval{\wd\iclr@boxA*\iclr@H/(\ht\iclr@boxA+\dp\iclr@boxA)}pt}%
  \setlength\iclr@wB{\fpeval{\wd\iclr@boxB*\iclr@H/(\ht\iclr@boxB+\dp\iclr@boxB)}pt}%
  \noindent\begin{minipage}{\iclr@wA}{\iclrnofloats\input{#1}}\end{minipage}\hfill
  \begin{minipage}{\iclr@wB}{\iclrnofloats\iclr@scaletabulars\input{#2}}\end{minipage}}
\makeatother

\makeatletter
\DeclareRobustCommand\onedot{\futurelet\@let@token\@onedot}
\def\@onedot{\ifx\@let@token.\else.\null\fi\xspace}
\def\eg{\emph{e.g}\onedot} 
\def\ie{\emph{i.e}\onedot}

\makeatother

\let\cite\citep

\definecolor{iclrblue}{rgb}{0.21,0.49,0.74}
\usepackage[breaklinks,colorlinks,allcolors=iclrblue]{hyperref}
\usepackage[capitalize]{cleveref}
\crefname{section}{Section}{Sections}
\Crefname{section}{Section}{Sections}
\crefname{figure}{Figure}{Figures}
\Crefname{figure}{Figure}{Figures}
\crefname{table}{Table}{Tables}
\Crefname{table}{Table}{Tables}
\crefname{equation}{Eq.}{Eqs.}
\Crefname{equation}{Equation}{Equations}

\title{Video Diffusion Avatars: Taming Video Diffusion Models into Controllable 3D Avatars}
\author{Simon Giebenhain$^{1}$ \quad Tobias Kirschstein$^{1}$ \quad Liam Schoneveld$^{2}$ \\
\textbf{Davide Davoli$^{3}$\thanks{Providing contracted services for Toyota.} \quad Zhe Chen$^{2}$ \quad Matthias Nie{\ss}ner$^{1}$} \\
$^{1}$Technical University of Munich \quad $^{2}$Woven by Toyota \quad $^{3}$Toyota Motor Europe \\
\url{https://simongiebenhain.github.io/video-diffusion-avatars/}
}

\begin{document}
\maketitle
\ifdefined\vdaarxiv\lhead{\vdaarxivheader}\rhead{\vdaarxivrheader}\fi
\begin{strip}
\vspace{-14pt}
\centering
\begin{overpic}[width=\linewidth]{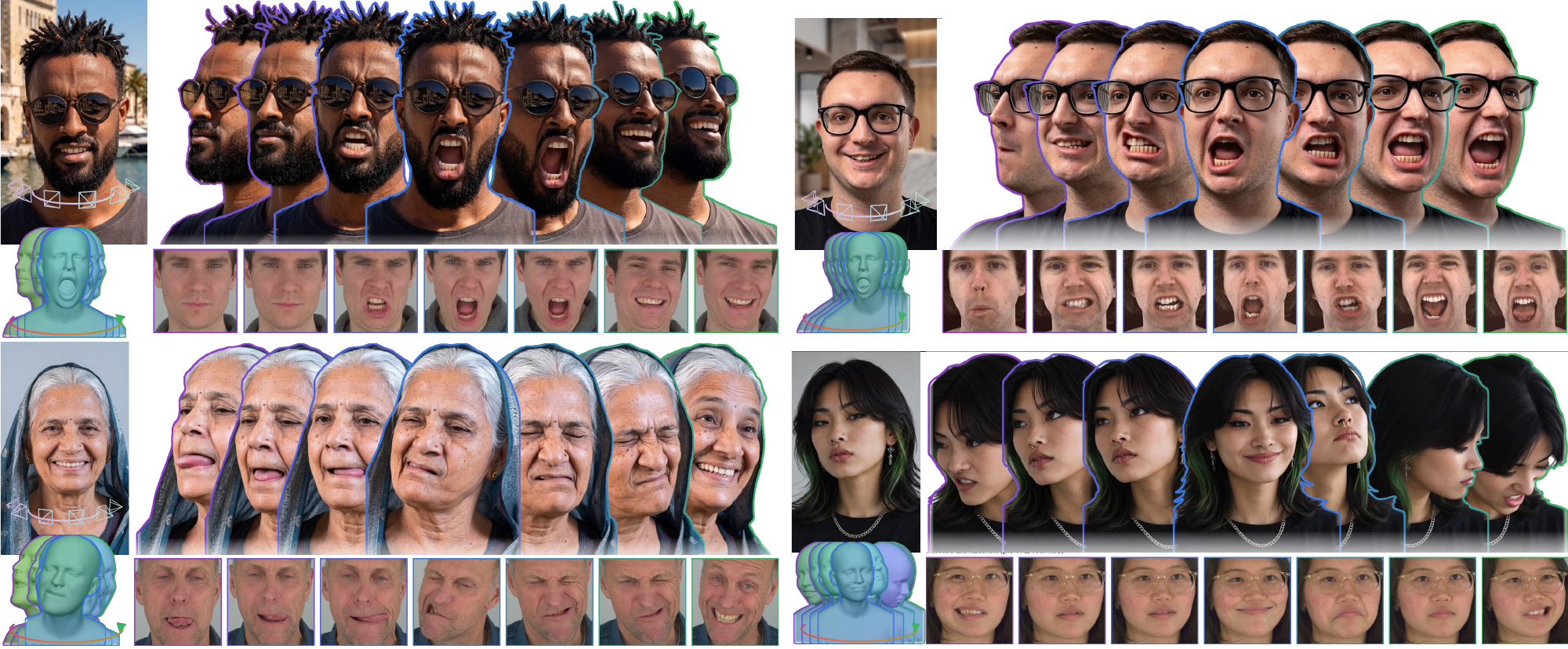}
\end{overpic}
\vspace{-14pt}
\captionof{figure}{
\textbf{Video Diffusion Avatars (VDAs)}: given a single image, VDAs produce photo-realistic renderings with
independent control over camera, upper-body pose, and facial expression, inheriting hair dynamics, shadows, reflections and refractions from its VDM prior.
}
\vspace{4pt}
\label{fig:teaser}
\end{strip}

\begin{abstract}
Video diffusion models (VDMs) trained on internet-scale data encode a rich
prior on appearance, motion, and physics that is difficult to replicate with
explicit 3D rendering and costly 3D data capture.
We introduce Video Diffusion Avatars (VDAs), a method that creates
photo-realistic, multi-view consistent avatars from a single image by
fine-tuning a VDM for precise control over camera viewpoint, facial
expression, and upper-body pose.
Departing from 3D morphable models (3DMMs), VDAs use a learned expression
space that lifts the expressiveness ceiling of 3DMMs, \eg for the tongue and
mouth interior. However, learned expression codes exhibit heavy viewpoint
bias, as our DISTRACT benchmark reveals. We show that this bias can be
removed through disentanglement, enabling their use as driving signals for
multi-view consistent avatars.
Camera and upper-body pose are controlled jointly via proxy-mesh renderings,
giving independent control over expression, pose, and viewpoint.
A camera-triplet training strategy makes the approach efficient: our models
train for 34 hours on a single A100 using only public multi-view data.
VDAs outperform state-of-the-art single-image 3D avatars on novel-view
reenactment, reducing face LPIPS by 29\% and improving perceptual quality
(JOD) by 66\% on BecomingLit.
A static 3DGS reconstruction probe further confirms that our generations are
the most multi-view consistent.
Built on a video prior, VDAs produce effects beyond the reach of explicit
renderers, including physically plausible hair dynamics, shadows,
reflections, and refractions.
\end{abstract}
\vspace{-10pt}

\section{Introduction}
\label{sec:intro}

Virtual avatars promise to reshape communication, gaming, and visual
effects. Many applications demand exact control over facial expression,
upper-body pose, and camera viewpoint, and the smallest artifacts read as
uncanny. In-the-wild use raises the bar further: accessories, complex
lighting, and hair dynamics must be handled convincingly from a single
reference image.\\
The dominant paradigm builds explicit volumetric avatars, typically
3DGS~\cite{kerbl20233dgs} or voxels~\cite{lombardi2021mixture,cao2022authentic}, with expressions driven by
3DMMs~\cite{blanz1999morphable,li2017flame}. This approach faces two
fundamental limits. First, no large-scale 3D training data exists, and, additionally,
effects such as shadows, reflections, refractions, and physically plausible
hair dynamics are hard to render explicitly. Second, 3DMM-driven expression
control suffers fidelity loss from 2D-to-3D registration. Alternatively, learned expression
spaces~\cite{drobyshev2022megaportraits,cheng2025wananimate,zhao2025xnemo}
avoid this ceiling but, as our proposed DISTRACT benchmark reveals, carry a heavy
viewpoint bias that must be addressed before they can drive multi-view
consistent avatars.\\
Video diffusion models (VDMs), trained on internet-scale data, offer a rich
prior on 3D geometry, photo-realistic appearance, motion, and physics.
However, existing VDM-based avatar methods either build 2D avatars without
camera control~\cite{tian2024emo,xie2024xportrait,guo2024liveportrait,zhao2025xnemo},
or use the VDM only to synthesize multi-view images that are then distilled
into a 3DGS avatar~\cite{taubner2025cap4d,mvp4d}, trading off rendering quality against speed.\\
We introduce Video Diffusion Avatars (\OURS), and depart from the two most common building blocks for virtual avatars: 3DMM-control and 3DGS rendering.
Following DH-STAR~\cite{zielonka2026digitalhumans}, we define an avatar as a model
that, conditioned on a single identity image, renders that person under any
requested facial expression, upper-body pose, and camera viewpoint. In our
approach the VDM \emph{is} the avatar: no 3D asset is ever built, and the
prior is fully preserved.
\OURS rests on three cornerstones.
\emph{Expression control:} we remove the viewpoint bias from learned
expression encoders with an explicit disentangler%
, obtaining higher expression fidelity
than FLAME-driven control.
\emph{Camera and pose control:} a proxy mesh combining a FLAME head with a
SAM-3D-Body~\cite{yang2026sam3dbody} torso jointly encodes camera viewpoint
and upper-body pose, while exposing separate control for both.
\emph{Training:} camera triplets from curated multi-view
data~\cite{kirschstein2023nersemble,martinez2024codecavatarstudio}, with
reference, driver and target from different cameras, teach multi-view
consistency in 34 hours on a single A100.\\
\OURS outperform all state-of-the-art single-image 3D avatars on
BecomingLit~\cite{schmidt2025becominglit},
Multiface~\cite{wuu2022multiface}, and VFHQ-test~\cite{xie2022vfhq},
improving LPIPS by 29\% and JOD by 66\% over the strongest baseline. A
3DGS optimization against our generations outperforms prior diffusion-based
avatars~\cite{taubner2025cap4d,mvp4d}, camera-controlled VDMs~\cite{facecam}, and 3DGS-based
avatars~\cite{kirschstein2026flexavatar,wu2026uika}.
Our contributions are:
\begin{enumerate}[label=(\alph*),nosep,leftmargin=*,labelsep=3pt,topsep=1pt]
\item \OURS replace the explicit 3D asset with a camera- and expression-controlled VDM, inheriting shadows, reflections, refractions, and hair dynamics from its prior.
\item We expose viewpoint bias in learned expression encoders with the
  DISTRACT benchmark, debias it, and obtain higher
  expression fidelity than FLAME-driven control.
\item Camera and pose conditioning via a Pl\"ucker-colored proxy mesh, that separates control over camera viewpoint from neck pose.
\item Affordable training strategy on public multi-view data using camera triplets in 34 hours on a single A100, which preserves performance on challenging in-the-wild inputs.
\end{enumerate}

\section{Related Work}
\label{sec:related}

\cparagraph{Facial expression spaces and 3D morphable models.}
3DMMs~\cite{li2017flame,blanz1999morphable} are the dominant expression parameterization, but
they frequently miss micro-expressions and lack a mouth interior, limiting the fidelity of any
system built on them. However, feed-forward 3DMM
regressors such as DECA~\cite{feng2021deca}, EMOCA~\cite{danecek2022emoca}, and Pix2NPHM~\cite{giebenhain2025pix2nphm},
and robust tracking~\cite{giebenhain2025pixel3dmm} are widely available and make
3DMM-based control practical for in-the-wild applications. Analogous body
models include SMPL~\cite{loper2015smpl} and
SAM-3D-Body~\cite{yang2026sam3dbody}.
Implicit neural head
representations~\cite{giebenhain2023nphm,giebenhain2024mononphm,zheng2022imface}
go beyond the linear 3DMM expression space. Alternatively, several methods learn an
expression latent directly from 2D data:
MegaPortraits~\cite{drobyshev2022megaportraits} relies on
cycle-consistency, while WAN-Animate~\cite{cheng2025wananimate} and
X-NeMo~\cite{zhao2025xnemo} learn expression representations end-to-end
for video-driven animation. In \cref{sec:benchmark} we compare the qualities
of such learned expression spaces and reveal a heavy viewpoint bias that
must be addressed for multi-view consistent avatars.

\cparagraph{Avatars.}
Most recent avatar systems use neural radiance fields (NeRFs)~\cite{mildenhall2020nerf} and 3DGS~\cite{kerbl20233dgs}, either for
generalizable single-image
creation~\cite{kirschstein2026flexavatar,he2025lam,wu2026uika,kirschstein2024gghead} or for
per-identity optimization from multi-view
captures~\cite{xu2024gaussianheadavatar,qian2024gaussianavatars,giebenhain2024npga}.
A separate line distills the outputs of diffusion models into an explicit
3D representation: CAP4D~\cite{taubner2025cap4d} and Joker~\cite{prinzler2025joker} condition an image-diffusion
model on 3DMM renderings and distill its multi-view images into a 3DGS/NeRF avatar, and MVP4D~\cite{mvp4d} uses a video-diffusion model
instead. These methods trade the photorealistic diffusion prior for an
explicit 3D asset. We refer to DH-STAR~\cite{zielonka2026digitalhumans} for a
broader survey.
\cparagraph{Video diffusion models for avatars.}
VDMs such as WAN-Animate-14B~\cite{wan2025wan,cheng2025wananimate}
synthesize temporally coherent, photorealistic human video at unprecedented
quality. Diffusion-based portrait-animation
methods~\cite{guo2024liveportrait,tian2024emo,xie2024xportrait,zhao2025xnemo,li2025personalive}
condition these models on signals ranging from sparse 2D keypoints to
dense 3DMM-derived renderings, but typically produce 2D avatars without
explicit camera control. DiffusionAvatars~\cite{kirschstein2024diffusionavatars} trains a person-specific image-diffusion model as the avatar.
FaceCam~\cite{facecam} adds camera control to a VDM, but lacks expression control.
DeX-Portrait~\cite{wu2025dexportrait} controls neck pose and expression,
but not the camera.
FactorPortrait~\cite{tang2025factorportrait} is concurrent work that
controls expression, head pose, and viewpoint in a VDM. However, it trains on
proprietary data using limited-quality 3DGS renderings as pseudo ground truth, and no code
or model is available.

\section{Video Diffusion Avatars (VDAs)}
\label{sec:method}

\begin{figure*}[t]
\centering
\includegraphics[width=\linewidth]{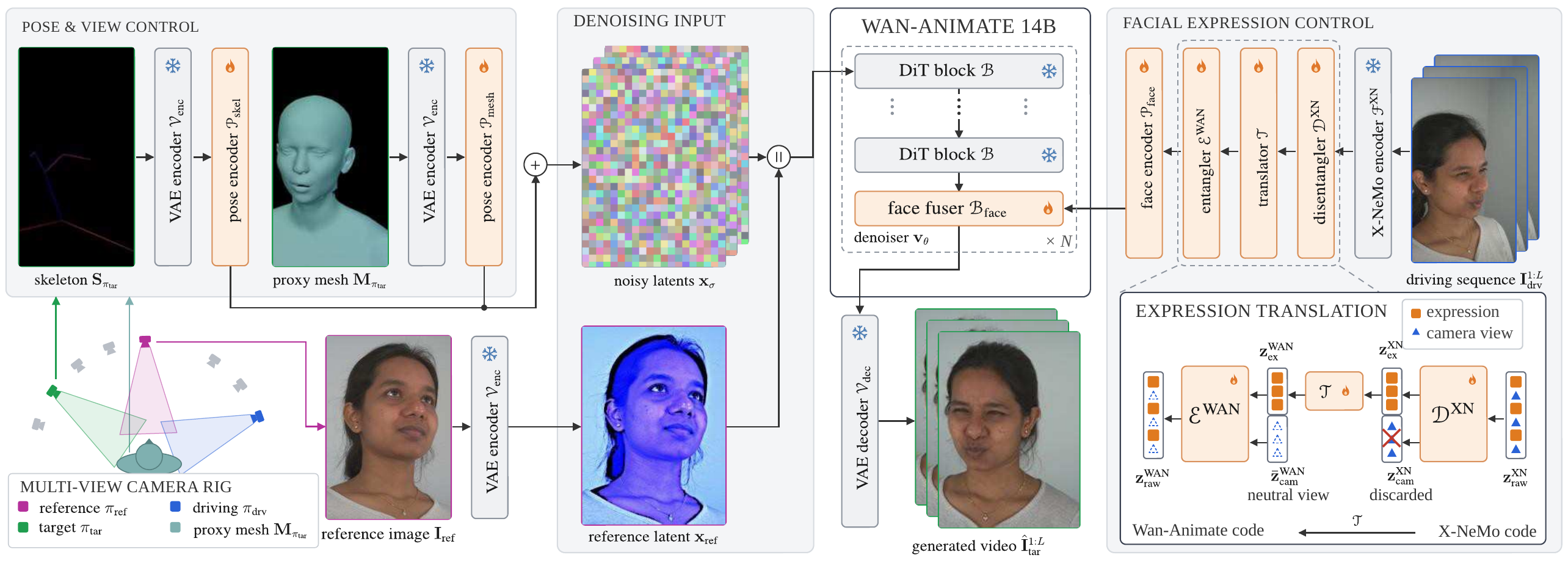}
\vspace{-14pt}
\caption{\textbf{Video Diffusion Avatar architecture}:
Camera, upper-body pose, and facial expression are injected as
independent control signals into WAN-Animate-14B.
Disentangled X-NeMo expression codes are translated into
WAN-Animate's native conditioning format with a viewpoint-neutral
camera code, while a proxy-mesh rendering encodes camera and pose.
Camera triplets from multi-view data teach multi-view consistency.}
\vspace{-10pt}
\label{fig:method}
\end{figure*}

Following DH-STAR~\cite{zielonka2026digitalhumans}, an avatar is a
function $\net{A}$ that renders a person, given an identity specification, under
control signals for expression, body pose, and camera viewpoint. Our avatar takes
a single reference image $\mathbf{I}_{\text{ref}}$ as identity input and
renders a video $\hat{\mathbf{I}}_{\text{tar}}^{1:L}$ conditioned on
per-frame disentangled expression codes $\mathbf{z}_{\text{ex}}^{1:L}$
(\cref{sec:benchmark}), and a proxy mesh
$\mathbf{M}_{\pi_{\text{tar}}}$ and skeleton
$\mathbf{S}_{\pi_{\text{tar}}}$ rendered from a target camera
$\pi_{\text{tar}}$, describing camera viewpoint and upper-body pose
(\cref{sec:poseconditioning}):
\begin{equation}
  \hat{\mathbf{I}}_{\text{tar}}^{1:L} = \net{A}\big(\mathbf{z}_{\text{ex}}^{1:L},\,
  \mathbf{M}_{\pi_{\text{tar}}}^{1:L},\,
  \mathbf{S}_{\pi_{\text{tar}}}^{1:L} \mid \mathbf{I}_{\text{ref}}\big).
\end{equation}
\Cref{fig:method} illustrates the architecture and how each signal enters
our fine-tuned WAN-Animate-14B backbone: expression through dedicated face-fuser blocks
(\cref{sec:arch}), camera and pose through the rendered proxy mesh and
skeleton (\cref{sec:poseconditioning}). Unlike a 3D avatar, which builds an
explicit 3D representation and renders it, $\net{A}$ synthesizes the video
directly, so multi-view consistency has to be learned rather than built in.
\Cref{sec:training} shows how camera triplets from public multi-view data
enable extremely efficient training (34 hours on a single GPU), while preserving the video prior.

\ificlr
\ificlr\subsection{Expression Control and the DISTRACT Benchmark}\else\section{Benchmark}\fi
\label{sec:benchmark}

3DMMs such as FLAME limit the detail and range of representable expressions,
motivating the use of learned expression encoders. A good encoder must
preserve expression while discarding identity and camera viewpoint. We
introduce the \textbf{DISTRACT} benchmark, which directly measures how
strongly an encoder's code is \emph{distracted} by these confounders.\\
DISTRACT evaluates two invariance axes on
NeRSemble~\cite{kirschstein2023nersemble}: \emph{viewpoint invariance}
(same identity and timestep, different camera) and \emph{identity
invariance} (same expression across different identities, verified by
manual inspection). Full details are in \cref{sec:supp-evalprotocol}.\\
Most encoders are heavily viewpoint-biased
(\cref{tab:distract,fig:distract}): X-NeMo reaches only $18\%$ Rank-1
on viewpoint invariance. Yet a small MLP probe on the \emph{same frozen}
latents raises this to $98\%$, confirming that expression information
survives but is entangled with viewpoint, thus motivating an explicit
disentanglement module.\\
\ificlr
\begin{table}[t]
\begin{figure}[t]
\centering
\includegraphics[width=\linewidth]{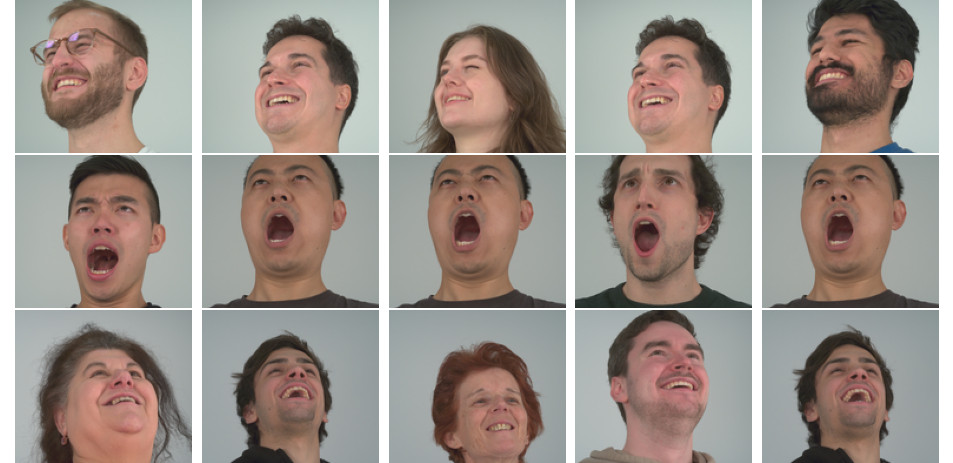}
\vspace{-14pt}
\caption{Nearest-neighbour retrieval in X-NeMo's expression space:
viewpoint dominates, yet expression information is retained.}
\label{fig:distract}
\end{figure}

\vspace{-10pt}
\end{table}
\else
\begin{table}[t]
\centering
\caption{DISTRACT results (\%) on raw codes and after a 2-layer MLP probe. Err-X: how often distractor type X outranks the true positive.}
\vspace{-4pt}
\label{tab:distract}
\footnotesize
\setlength{\tabcolsep}{1.0pt}
\setlength{\fboxsep}{0.6pt}
\begin{tabular}{clcccc}
\toprule
 & & \multicolumn{2}{c}{Latents} & \multicolumn{2}{c}{MLP Probe} \\
\cmidrule(lr){3-4} \cmidrule(lr){5-6}
 & Encoder & R-1$\uparrow$ & \shortstack{Err\\A/B/C/D$\downarrow$} & R-1$\uparrow$ & \shortstack{Err\\A/B/C/D$\downarrow$} \\
\midrule
\multirow{4}{*}{\rotatebox[origin=c]{90}{\tiny View Inv.}} & DECA~\cite{feng2021deca}               & \cellcolor{tabfirst}{61} & \colorbox{tabfirst}{33}/\colorbox{tabfirst}{18}/\colorbox{tabfirst}{18}/\colorbox{tabfirst}{11} & {85} & {12}/{06}/\colorbox{tabthird}{01}/\colorbox{tabsecond}{00} \\
 & PD-FGC~\cite{wang2023pdfgc}            & {08} & {90}/{42}/{79}/{43} & \cellcolor{tabthird}{90} & \colorbox{tabthird}{08}/\colorbox{tabthird}{03}/\colorbox{tabsecond}{00}/\colorbox{tabthird}{01} \\
 & WAN-Animate~\cite{cheng2025wananimate} & \cellcolor{tabsecond}{21} & \colorbox{tabsecond}{72}/\colorbox{tabsecond}{34}/\colorbox{tabsecond}{59}/\colorbox{tabsecond}{33} & \cellcolor{tabsecond}{94} & \colorbox{tabsecond}{05}/\colorbox{tabsecond}{01}/\colorbox{tabthird}{01}/\colorbox{tabsecond}{00} \\
 & X-NeMo~\cite{zhao2025xnemo}            & \cellcolor{tabthird}{18} & \colorbox{tabthird}{78}/\colorbox{tabthird}{35}/\colorbox{tabthird}{70}/\colorbox{tabthird}{34} & \cellcolor{tabfirst}{98} & \colorbox{tabfirst}{02}/\colorbox{tabfirst}{00}/\colorbox{tabfirst}{00}/\colorbox{tabfirst}{00} \\
\midrule
\multirow{4}{*}{\rotatebox[origin=c]{90}{\tiny Identity Inv.}} & DECA~\cite{feng2021deca}               & \cellcolor{tabfirst}{34} & \colorbox{tabfirst}{51}/\colorbox{tabsecond}{37}/\colorbox{tabfirst}{25}/\colorbox{tabfirst}{24} & {32} & {58}/{51}/{24}/\colorbox{tabthird}{16} \\
 & PD-FGC~\cite{wang2023pdfgc}            & \cellcolor{tabthird}{07} & {91}/\colorbox{tabthird}{48}/{81}/{43} & \cellcolor{tabthird}{46} & \colorbox{tabthird}{44}/\colorbox{tabthird}{40}/\colorbox{tabsecond}{10}/{18} \\
 & WAN-Animate~\cite{cheng2025wananimate} & \cellcolor{tabsecond}{14} & \colorbox{tabsecond}{74}/\colorbox{tabfirst}{32}/\colorbox{tabsecond}{62}/\colorbox{tabsecond}{26} & \cellcolor{tabsecond}{66} & \colorbox{tabsecond}{22}/\colorbox{tabsecond}{23}/\colorbox{tabthird}{10}/\colorbox{tabsecond}{06} \\
 & X-NeMo~\cite{zhao2025xnemo}            & \cellcolor{tabsecond}{14} & \colorbox{tabthird}{78}/\colorbox{tabsecond}{37}/\colorbox{tabthird}{68}/\colorbox{tabthird}{37} & \cellcolor{tabfirst}{76} & \colorbox{tabfirst}{18}/\colorbox{tabfirst}{17}/\colorbox{tabfirst}{04}/\colorbox{tabfirst}{05} \\
\bottomrule
\end{tabular}
\end{table}

\begin{figure}[t]
\centering
\includegraphics[width=\linewidth]{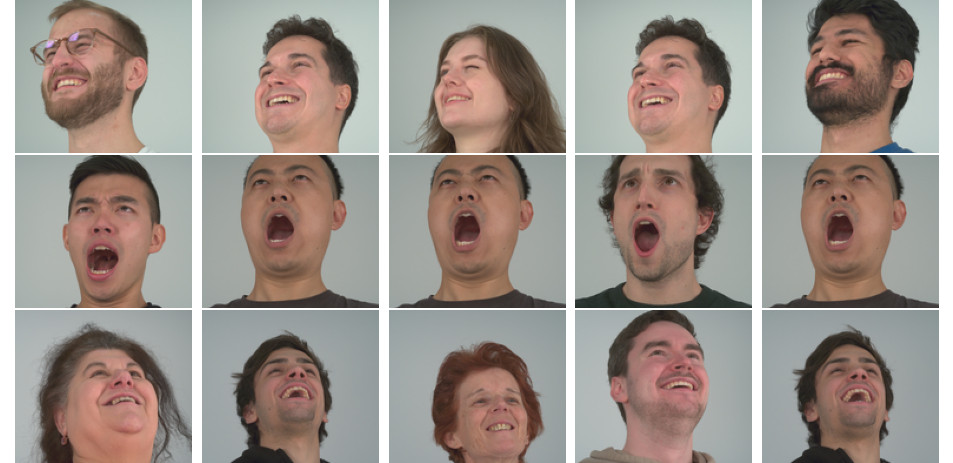}
\vspace{-14pt}
\caption{Nearest-neighbour retrieval in X-NeMo's expression space:
viewpoint dominates, yet expression information is retained.}
\label{fig:distract}
\end{figure}

\fi
Given an image $\mathbf{I}$, a learned expression encoder produces an
entangled code $\mathbf{z}_{\text{raw}} \in \mathbb{R}^d$. Our
\emph{disentangler} $\net{D}$ splits it into an expression
latent $\mathbf{z}_{\text{ex}}$ and a camera latent $\mathbf{z}_{\text{cam}}$, and an
\emph{entangler} $\net{E}$ reconstructs the original:
\begin{equation}
  \net{D}\big(\mathbf{z}_{\text{raw}}\big) = \big(\mathbf{z}_{\text{ex}},\, \mathbf{z}_{\text{cam}}\big),
  \qquad
  \net{E}\big(\mathbf{z}_{\text{ex}},\, \mathbf{z}_{\text{cam}}\big) = \hat{\mathbf{z}}_{\text{raw}} \approx \mathbf{z}_{\text{raw}},
\end{equation}
both implemented as dedicated MLP heads. The architecture is
encoder-agnostic; we train separate instances for X-NeMo and
WAN-Animate codes. Training uses a reconstruction
loss, a supervised contrastive loss~\cite{khosla2020supcon} on
$\mathbf{z}_{\text{ex}}$, and a cycle-consistency loss; details are in
\cref{sec:supp-disent}.

\ificlr\else

The cycle-consistency loss encourages a clean and stable factorization. Given two entangled codes $\mathbf{a}, \mathbf{b} \in \mathbb{R}^d$, we first disentangle them as
\begin{equation}
\net{D}^{\text{XN}}(\mathbf{a}) = (\mathbf{a}_{\text{ex}}, \mathbf{a}_{\text{cam}}), \quad \net{D}^{\text{XN}}(\mathbf{b}) = (\mathbf{b}_{\text{ex}}, \mathbf{b}_{\text{cam}}).
\end{equation}
We then swap the camera latents and pass the resulting codes through the entangler and disentangler:
\begin{equation}
(\hat{\mathbf{a}}_{\text{ex}}, \hat{\mathbf{b}}_{\text{cam}}) = \net{D}^{\text{XN}} \!\left( \net{E}^{\text{XN}} (\mathbf{a}_{\text{ex}}, \mathbf{b}_{\text{cam}}) \right), \qquad
(\hat{\mathbf{b}}_{\text{ex}}, \hat{\mathbf{a}}_{\text{cam}}) = \net{D}^{\text{XN}} \!\left( \net{E}^{\text{XN}} (\mathbf{b}_{\text{ex}}, \mathbf{a}_{\text{cam}}) \right).
\end{equation}
The cycle-consistency loss enforces recovery of the original factors after this swap-and-reconstruct operation:
\begin{equation}
\mathcal{L}_{\text{cycle}} = \sum_{k \in \{\text{ex},\text{cam}\}} \left( \|\hat{\mathbf{a}}_{k}-\mathbf{a}_{k}\| + \|\hat{\mathbf{b}}_{k}-\mathbf{b}_{k}\| \right).
\end{equation}

We additionally apply a supervised contrastive loss to the expression latent $\zc{XN}{ex}$. Samples sharing the same expression label are treated as positive pairs, while samples with different expression labels are treated as negatives. The objective encourages expression latents from the same class to be close in the embedding space and those from different classes to be far apart: \begin{equation} \mathcal{L}_{\mathrm{con}} = -\sum\limits_{p \in \mathcal{P}} \log \frac{  \exp\!\bigl( \mathrm{sim}(\mathbf{z}^{i}_{\text{ex}}, \mathbf{z}^{p}_{\text{ex}})/\tau \bigr) }{ \sum\limits_{a \in \left\{ \mathcal{P} \cup \mathcal{N} \right\} } \exp\!\bigl( \mathrm{sim}(\mathbf{z}^{i}_{\text{ex}}, \mathbf{z}^{a}_{\text{ex}})/\tau \bigr) }, \end{equation} where $\mathcal{P}$ and $\mathcal{N}$ denote the positive and negative candidate samples for anchor $i$, respectively.

\fi

\fi

\subsection{Latent Expression Injection}
\label{sec:arch}
We drive \OURS using X-NeMo expression codes, which outperformed the WAN-Animate codes on DISTRACT after the MLP probe. Concretely, we replace the view-biased codes $\zc{WAN}{raw}$ with our disentangled counterpart $\zc{XN}{ex}$.
Since the two latent spaces are unrelated, and we aim to minimally disturb the pretrained weight space, we bridge this gap using latent translation, depicted in \cref{fig:method}. An MLP \emph{translator} $\net{T}$
maps the disentangled X-NeMo expression latent into WAN-Animate's
disentangled expression space, and the \emph{entangler}
$\net{E}^{\text{WAN}}$ combines it with a viewpoint-neutral camera code
$\bar{\mathbf{z}}^{\text{WAN}}_{\text{cam}}$, the mean of the camera latents
of $\net{D}^{\text{WAN}}$ over the training set:
\begin{equation}
  \zc{WAN}{ex} = \net{T}\big(\zc{XN}{ex}\big), \qquad
  \zc{WAN}{raw} = \net{E}^{\text{WAN}}\big(\zc{WAN}{ex},\,
  \bar{\mathbf{z}}^{\text{WAN}}_{\text{cam}}\big).
\end{equation}
The result is a code in WAN-Animate's native format, now stripped of
viewpoint bias. WAN-Animate's face encoder $\net{P}_{\text{face}}$ maps it to
face tokens, which enter the denoiser $\mathbf{v}_\theta$ through face-fuser
blocks $\net{B}_{\text{face}}$ interleaved with its DiT blocks $\net{B}$. Since the DiT has not seen view-neutralized codes during pretraining, and our disentanglement is not trained with the DiT in-the-loop, we fine-tune
$\net{T}$, $\net{E}^{\text{WAN}}$, $\net{D}^{\text{XN}}$, $\net{P}_{\text{face}}$,
$\net{B}_{\text{face}}$, and the pose encoders (\cref{sec:poseconditioning}), while we keep $\net{B}$ frozen to preserve as much prior as possible.

\subsection{Pose Conditioning}
\label{sec:poseconditioning}

WAN-Animate relies on 2D skeleton prediction from
ViTPose~\cite{vitpose} for pose control, which cannot be transferred to
novel viewpoints for novel-view synthesis. For 3D-consistent skeleton
control, we replace the 2D detections with skeletons derived from
SAM-3D-Body~\cite{yang2026sam3dbody}, a single-image 3D body estimator.
To reduce the noise of SAM-3D-Body's per-frame pose estimates, we
introduce \emph{pelvis-anchored kinematic fitting}
(\cref{sec:supp-seated}), which exploits the seated capture setup and
assumes a constant pelvis position to remove jitter and drift. Importantly, we retain WAN-Animate's skeleton conditioning format for better prior preservation. We note that at inference, the seated assumption can be relaxed to plain temporal smoothing.\\
However, we find that skeleton conditioning alone is too ambiguous for precise camera control and that \OURS
benefits from a more precise and correlated signal. We therefore render a
proxy mesh $\mathbf{M}$ combining a FLAME head with a SAM-3D-Body torso, that
jointly encodes camera viewpoint and body/neck pose. We color each pixel
by the direction of its camera ray, \ie the directional part of its
Pl\"ucker coordinates, blended with Lambertian shading. Crucially, head
pixels express this direction in the head's local frame and torso pixels
in a pelvis-aligned frame (\cref{sec:supp-plucker}). The color contrast
between head and torso thus directly encodes neck pose, keeping camera
viewpoint and neck rotation separable. Rendering from the target camera
$\pi_{\text{tar}}$ yields $\mathbf{S}_{\pi_{\text{tar}}}$ and
$\mathbf{M}_{\pi_{\text{tar}}}$, which are encoded by
$\net{P}_{\text{skel}}$ and $\net{P}_{\text{mesh}}$ (the latter
zero-initialized) and added to the noisy video latents.

\subsection{Training with Camera Triplets}
\label{sec:training}

We train \OURS on public multi-view video datasets of human faces.
Each training sample is a \emph{camera triplet}
$(\pi_{\text{ref}}, \pi_{\text{drv}}, \pi_{\text{tar}})$ of three
\emph{different} cameras of the same multi-view video clip. 
The reference image
$\mathbf{I}_{\text{ref}}$ is seen from $\pi_{\text{ref}}$, the driving clip
$\mathbf{I}_{\text{drv}}^{1:L}$ from $\pi_{\text{drv}}$, and the target clip
$\mathbf{I}_{\text{tar}}^{1:L}$ from $\pi_{\text{tar}}$. 
The expression codes are extracted from the driving video with the frozen
X-NeMo image encoder $\net{F}^{\text{XN}}$ and our disentangler:
\begin{equation}
  \mathbf{z}_{\text{ex}}^{1:L} = \net{D}^{\text{XN}}\big(\net{F}^{\text{XN}}(\mathbf{I}_{\text{drv}}^{1:L})\big).
\end{equation}
Similarly, the proxy mesh is estimated from $\mathbf{I}_{\text{drv}}^{1:L}$, with the exception of FLAME's identity parameters which come from $\pi_{\text{ref}}$, and rendered from $\pi_{\text{tar}}$ to obtain $\mathbf{M}_{\pi_{\text{tar}}}^{1:L}$ and $\mathbf{S}_{\pi_{\text{tar}}}^{1:L}$.
Because the $\pi_{drv}$ and $\pi_{tar}$ differ, viewpoint
information inside $\mathbf{z}_{\text{ex}}$ is actively misleading and
training suppresses it (\cref{sec:ablations}).\\
We fine-tune with the rectified-flow matching loss. Given the target
latent $\mathbf{x}_0 = \net{V}_{\text{enc}}(\mathbf{I}_{\text{tar}}^{1:L})$
and Gaussian noise $\boldsymbol{\epsilon}$, the noised latent is
$\mathbf{x}_\sigma = (1 - \sigma)\,\mathbf{x}_0 + \sigma\,\boldsymbol{\epsilon}$,
and the denoiser $\mathbf{v}_\theta$ predicts the velocity
$\mathbf{v} = \boldsymbol{\epsilon} - \mathbf{x}_0$:
\begin{equation}
  \mathcal{L} = \mathbb{E}\bigl[w(\sigma)\,\bigl\lVert
    \mathbf{v}_\theta\bigl(\mathbf{x}_\sigma,\, \sigma;\, \mathbf{z}_{\text{ex}}^{1:L},\,
    \mathbf{M}_{\pi_{\text{tar}}}^{1:L},\, \mathbf{S}_{\pi_{\text{tar}}}^{1:L},\, \mathbf{x}_{\text{ref}}\bigr)
    - \mathbf{v} \bigr\rVert^2\bigr],
\end{equation}
where $\mathbf{x}_{\text{ref}} = \net{V}_{\text{enc}}(\mathbf{I}_{\text{ref}})$
is concatenated to the noisy latents and excluded from the loss, and
$w(\sigma)$ is a bell-shaped weighting peaking at $\sigma = 0.5$. At
inference, $\net{A}$ integrates $\mathbf{v}_\theta$ from $\sigma = 1$ to $0$
and decodes the result with the VAE decoder $\net{V}_{\text{dec}}$.
No auxiliary losses are used. For more training details we refer to
\cref{sec:supp-impl}.

\section{Experiments}
\label{sec:experiments}

\ificlr
\begin{table*}[t]
\centering
\caption{Quantitative results (BecomingLit/Multiface per cell). MMDM is
CAP4D's diffusion component. Cross-reenactment on BecomingLit only; colours
rank within each dataset.}
\vspace{-4pt}
\label{tab:selfreenact}
\footnotesize
\setlength{\tabcolsep}{1.6pt}
\setlength{\fboxsep}{0.7pt}
\begin{tabular}{lccccccccc}
\toprule
 & \multicolumn{6}{c}{Self-reenactment ($n_{\text{BecomingLit}}$=25 / $n_{\text{Multiface}}$=22)} & \multicolumn{3}{c}{Cross-reenact.\ ($n$=13)} \\
\cmidrule(lr){2-7} \cmidrule(lr){8-10}
Method & PSNR~$\uparrow$ & LPIPS~$\downarrow$ & JOD~$\uparrow$ & ID~$\uparrow$ & Expr~$\uparrow$ & AKD~$\downarrow$ & ID~$\uparrow$ & Expr~$\uparrow$ & AKD~$\downarrow$ \\
\midrule
LAM        & 18.95/19.83 & 0.408/0.415 & 2.00/1.92 & 0.295/0.334 & 0.670/0.528 & 0.103/0.114 & 0.314 & 0.623 & 0.091 \\
FlexAvatar & 20.55/19.45 & 0.327/0.364 & 2.28/2.33 & \colorbox{tabsecond}{0.624}/\colorbox{tabsecond}{0.645} & \colorbox{tabsecond}{0.890}/\colorbox{tabsecond}{0.850} & 0.086/0.092 & \colorbox{tabsecond}{0.570} & 0.700 & 0.130 \\
UIKA       & 20.37/20.76 & 0.327/0.332 & \colorbox{tabthird}{2.34}/\colorbox{tabthird}{2.41} & 0.532/0.571 & \colorbox{tabthird}{0.865}/0.805 & 0.076/0.072 & 0.493 & \colorbox{tabthird}{0.775} & 0.084 \\
\midrule
MMDM       & \colorbox{tabsecond}{20.85}/21.02 & \colorbox{tabthird}{0.283}/\colorbox{tabsecond}{0.270} & 1.56/1.35 & 0.555/\colorbox{tabthird}{0.608} & 0.853/\colorbox{tabthird}{0.840} & \colorbox{tabsecond}{0.058}/0.063 & 0.522 & \colorbox{tabsecond}{0.780} & \colorbox{tabsecond}{0.069} \\
MVP4D      & 20.16/\colorbox{tabsecond}{21.64} & \colorbox{tabsecond}{0.282}/\colorbox{tabthird}{0.273} & 1.85/1.64 & 0.490/0.531 & 0.831/0.828 & 0.063/\colorbox{tabthird}{0.060} & 0.480 & 0.755 & 0.074 \\
CAP4D      & \colorbox{tabthird}{20.73}/\colorbox{tabthird}{21.40} & 0.322/0.303 & \colorbox{tabsecond}{2.40}/\colorbox{tabsecond}{2.49} & \colorbox{tabthird}{0.562}/\colorbox{tabthird}{0.608} & 0.852/0.826 & \colorbox{tabthird}{0.059}/\colorbox{tabsecond}{0.059} & \colorbox{tabthird}{0.529} & 0.770 & \colorbox{tabthird}{0.070} \\
\midrule
Ours       & \colorbox{tabfirst}{22.43}/\colorbox{tabfirst}{22.56} & \colorbox{tabfirst}{0.202}/\colorbox{tabfirst}{0.201} & \colorbox{tabfirst}{3.99}/\colorbox{tabfirst}{3.49} & \colorbox{tabfirst}{0.699}/\colorbox{tabfirst}{0.719} & \colorbox{tabfirst}{0.920}/\colorbox{tabfirst}{0.862} & \colorbox{tabfirst}{0.055}/\colorbox{tabfirst}{0.057} & \colorbox{tabfirst}{0.623} & \colorbox{tabfirst}{0.808} & \colorbox{tabfirst}{0.066} \\
\bottomrule
\end{tabular}
\vspace{-10pt}
\end{table*}

\fi

\ificlr\cparagraph{Implementation details.}\else\subsection{Implementation Details}\fi
We fine-tune WAN-Animate-14B on curated multi-view data from
NeRSemble~\cite{kirschstein2023nersemble} and
AVA-256~\cite{martinez2024codecavatarstudio}.
Training takes \textbf{34 hours on a single A100 GPU}.
The proxy-mesh head comes from monocular
pixel3DMM~\cite{giebenhain2025pixel3dmm} FLAME tracking; the body and
neck from SAM-3D~\cite{yang2026sam3dbody}.
At inference we perform 20 denoising steps.
Full hyperparameters are in \cref{sec:supp-impl}.

\ificlr\subsection{Evaluation Setup}\fi

\ificlr\cparagraph{Baselines.}\else\subsection{Baselines}\fi

We compare against the single-image 3DGS avatars
LAM~\cite{he2025lam}, FlexAvatar~\cite{kirschstein2026flexavatar}, and
UIKA~\cite{wu2026uika}, and against CAP4D~\cite{taubner2025cap4d} and
MVP4D~\cite{mvp4d}, which distill FLAME-conditioned image- and
video-diffusion outputs, respectively, into a 3DGS avatar. For CAP4D, we
report both the full pipeline (\textbf{CAP4D}) and its diffusion model
(\textbf{MMDM}). For computational reasons, we evaluate only
MVP4D's video-diffusion model (MMVDM, labeled \textbf{MVP4D} in tables and
figures). The camera-controlled VDM
FaceCam~\cite{facecam} lacks expression control, so we include it only in
the multi-view consistency probe. Details in \cref{sec:supp-baselines}.

\ificlr\cparagraph{Novel-View Reenactment Task.}\else\subsection{Evaluation Protocol}\fi
\label{sec:evalprotocol}
Our evaluation must simultaneously measure how well an avatar preserves
identity and expression \emph{and} how multi-view consistent it is.
We therefore make the protocol as viewpoint-challenging as possible: each
evaluation triplet consists of a single reference image
$\mathbf{I}_{\text{ref}}$, a driving video, and a target video from
\emph{different} camera viewpoints, similar to the camera-triplet setup
of training (\cref{sec:training}).
We evaluate on 
BecomingLit~\cite{schmidt2025becominglit} (25 self-reenactment and 13 cross-reenactment clips) and Multiface~\cite{wuu2022multiface} (22 self-reenactments), since no unified
cross-method split exists for NeRSemble or AVA-256. Those held-out datasets
eliminate training-data leakage for all baselines. Details are in \cref{sec:supp-evalprotocol}.

\cparagraph{3DGS Distillation.}
To explicitly measure multi-view consistency, we additionally fit a static
3DGS scene from each diffusion-based method's generations along a closed ellipse camera
trajectory with a frozen expression, and evaluate the reconstruction. We
choose the static setting because dynamic 3DGS optimization is far less
standardized and can itself introduce fitting flaws that would confound the
comparison. We render the fit at the 16 real rig cameras, none of which
is part of the generated trajectory, and compare against the captured
images. Details are in \cref{sec:supp-evalprotocol}.
\ificlr
Metric definitions are given in \cref{sec:metrics}.
\else

\subsection{Metrics}
\label{sec:metrics}

We report PSNR and LPIPS over a face-only crop. We also report JOD, a perceptual
video-quality metric computed via
pyfvvdp~\cite{mantiuk2021fovvideovdp} that captures temporal, video-level
quality beyond per-frame image metrics. We further report
face-specific semantic metrics: identity preservation as
ArcFace~\cite{deng2019arcface} cosine similarity to the ground-truth
target, and expression similarity as the cosine similarity between the
DECA~\cite{feng2021deca} FLAME expression codes of the generated and
ground-truth frames, averaged over the clip. We deliberately score
expression in a \emph{third-party} space rather than in
X-NeMo's~\cite{zhao2025xnemo} own motion embedding: our method is driven by
X-NeMo codes, so a cosine similarity in that same space would be a metric we
have a stake in. For pose accuracy we report
keypoint distance normalized by inter-ocular distance (AKD/IOD), and in the
ablations additionally APD, the average rotational distance (in degrees)
between the head poses estimated with the FAN3D landmark detector on the
generated and ground-truth frames. 

To make these and other
pixel-aligned metrics meaningful, we align the 3DMM present in every
compared model to the ground-truth 3DMM before computing them, removing an
otherwise-unresolved ambiguity. On the multi-view benchmarks this alignment
is possible because a ground-truth target view exists. On monocular
VFHQ-test it does not, so there we follow the standard protocol for that
dataset and align on detected facial keypoints instead, landmark-cropping
every method's output to a common head scale and centre before scoring.
That alignment is what makes the pixel metrics comparable across methods
whose outputs live in different frames, including methods that render
into the reference frame rather than reprojecting to a target camera.
Multi-view consistency is measured
separately with the 3DGS probe (\cref{sec:novelview}). Cross-reenactment has no ground-truth target image, so the
pixel metrics do not apply there. Identity is instead measured against the
reference, and expression and pose against the driver's own target-view
video --- the driver has the correct expression and pose for that camera,
only the wrong identity, which is exactly what those two metrics need.

\fi

\ificlr\subsection{Results}\fi

Human perception of faces is highly sensitive to temporal artifacts such
as flicker, identity drift, and unnatural motion, which single frames
cannot convey. We therefore strongly encourage viewing the supplementary
video alongside the figures below.

\begin{figure*}[t]
\centering
\includegraphics[width=\linewidth]{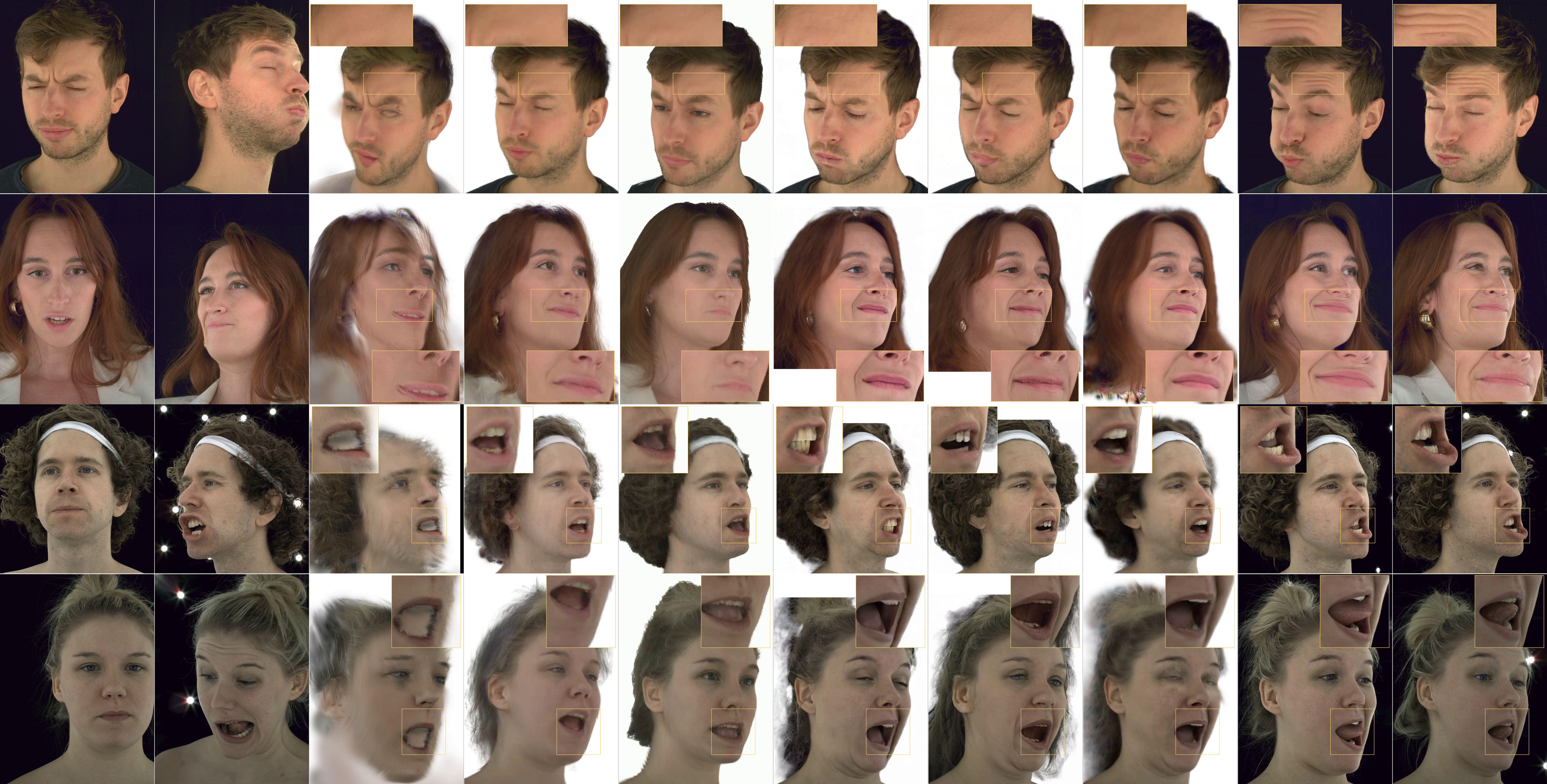}\\[1pt]
\vspace{-4pt}
\makebox[\linewidth][l]{\tiny
\makebox[0.09972\linewidth]{Reference}%
\hspace{0.00030\linewidth}%
\makebox[0.09972\linewidth]{Driving}%
\hspace{0.00030\linewidth}%
\makebox[0.09972\linewidth]{LAM}%
\hspace{0.00030\linewidth}%
\makebox[0.09972\linewidth]{FlexAvatar}%
\hspace{0.00030\linewidth}%
\makebox[0.09972\linewidth]{UIKA}%
\hspace{0.00030\linewidth}%
\makebox[0.09972\linewidth]{MMDM}%
\hspace{0.00030\linewidth}%
\makebox[0.09972\linewidth]{MVP4D}%
\hspace{0.00030\linewidth}%
\makebox[0.09972\linewidth]{CAP4D}%
\hspace{0.00030\linewidth}%
\makebox[0.09972\linewidth]{Ours}%
\hspace{0.00030\linewidth}%
\makebox[0.09972\linewidth]{Ground truth}%
}
\vspace{-14pt}
\caption{Self-reenactment on BecomingLit and Multiface.}
\vspace{-10pt}
\label{fig:selfreenact}
\end{figure*}

\ificlr\cparagraph{Self-reenactment.}\else\subsection{Self-Reenactment Results}\fi
\OURS outperform every baseline on every metric of both datasets
(\cref{tab:selfreenact}), as well as on VFHQ-test~\cite{xie2022vfhq}
(\cref{tab:vfhq}\ificlr, \cref{fig:selfreenact-vfhq} in the appendix\fi),
an in-the-wild single-view benchmark testing generalization beyond the
multi-view training distribution. Qualitatively (\cref{fig:selfreenact}),
our results retain sharper identity detail and transfer subtle
expressions and micro-expressions, as well as rare expressions outside
the FLAME space, which limits the baselines.

\ificlr\else

\fi

\ificlr\else
\begin{table}[t]
\centering
\caption{Quantitative results on the VFHQ-test~\cite{xie2022vfhq} dataset.}
\vspace{-5pt}
\label{tab:vfhq}
\footnotesize
\setlength{\tabcolsep}{1.5pt}
\begin{tabular}{lcccccc}
\toprule
Method & PSNR~$\uparrow$ & LPIPS~$\downarrow$ & ID~$\uparrow$ & Expr~$\uparrow$ & JOD~$\uparrow$ & AKD~$\downarrow$ \\
\midrule
LAM\tabcite{he2025lam}                        & \cellcolor{tabthird}20.4 & 0.050 & 0.605 & 0.791 & 3.0 & 0.037 \\
FlexAvatar\tabcite{kirschstein2026flexavatar} & \cellcolor{tabsecond}22.2 & \cellcolor{tabthird}0.043 & \cellcolor{tabsecond}0.804 & \cellcolor{tabthird}0.835 & \cellcolor{tabthird}3.3 & 0.038 \\
UIKA\tabcite{wu2026uika}                      & 19.8 & 0.053 & 0.736 & \cellcolor{tabsecond}0.842 & 3.0 & 0.043 \\
MMDM\tabcite{taubner2025cap4d}                & \cellcolor{tabsecond}22.2 & \cellcolor{tabsecond}0.040 & \cellcolor{tabthird}0.793 & 0.834 & \cellcolor{tabthird}3.3 & \cellcolor{tabthird}0.034 \\
MVP4D\tabcite{mvp4d}                          & \cellcolor{tabsecond}22.2 & \cellcolor{tabsecond}0.040 & 0.749 & 0.824 & \cellcolor{tabsecond}3.4 & \cellcolor{tabsecond}0.031 \\
Ours                                          & \cellcolor{tabfirst}24.9 & \cellcolor{tabfirst}0.024 & \cellcolor{tabfirst}0.852 & \cellcolor{tabfirst}0.931 & \cellcolor{tabfirst}5.0 & \cellcolor{tabfirst}0.028 \\
\bottomrule
\end{tabular}
\end{table}

\fi

\ificlr
\begin{table}[t]
\begin{table}[t]
\centering
\caption{Quantitative results on the VFHQ-test~\cite{xie2022vfhq} dataset.}
\vspace{-5pt}
\label{tab:vfhq}
\footnotesize
\setlength{\tabcolsep}{1.5pt}
\begin{tabular}{lcccccc}
\toprule
Method & PSNR~$\uparrow$ & LPIPS~$\downarrow$ & ID~$\uparrow$ & Expr~$\uparrow$ & JOD~$\uparrow$ & AKD~$\downarrow$ \\
\midrule
LAM\tabcite{he2025lam}                        & \cellcolor{tabthird}20.4 & 0.050 & 0.605 & 0.791 & 3.0 & 0.037 \\
FlexAvatar\tabcite{kirschstein2026flexavatar} & \cellcolor{tabsecond}22.2 & \cellcolor{tabthird}0.043 & \cellcolor{tabsecond}0.804 & \cellcolor{tabthird}0.835 & \cellcolor{tabthird}3.3 & 0.038 \\
UIKA\tabcite{wu2026uika}                      & 19.8 & 0.053 & 0.736 & \cellcolor{tabsecond}0.842 & 3.0 & 0.043 \\
MMDM\tabcite{taubner2025cap4d}                & \cellcolor{tabsecond}22.2 & \cellcolor{tabsecond}0.040 & \cellcolor{tabthird}0.793 & 0.834 & \cellcolor{tabthird}3.3 & \cellcolor{tabthird}0.034 \\
MVP4D\tabcite{mvp4d}                          & \cellcolor{tabsecond}22.2 & \cellcolor{tabsecond}0.040 & 0.749 & 0.824 & \cellcolor{tabsecond}3.4 & \cellcolor{tabsecond}0.031 \\
Ours                                          & \cellcolor{tabfirst}24.9 & \cellcolor{tabfirst}0.024 & \cellcolor{tabfirst}0.852 & \cellcolor{tabfirst}0.931 & \cellcolor{tabfirst}5.0 & \cellcolor{tabfirst}0.028 \\
\bottomrule
\end{tabular}
\end{table}

\end{table}
\fi

\ificlr\else
\begin{figure}[t]
\centering
\includegraphics[width=\linewidth]{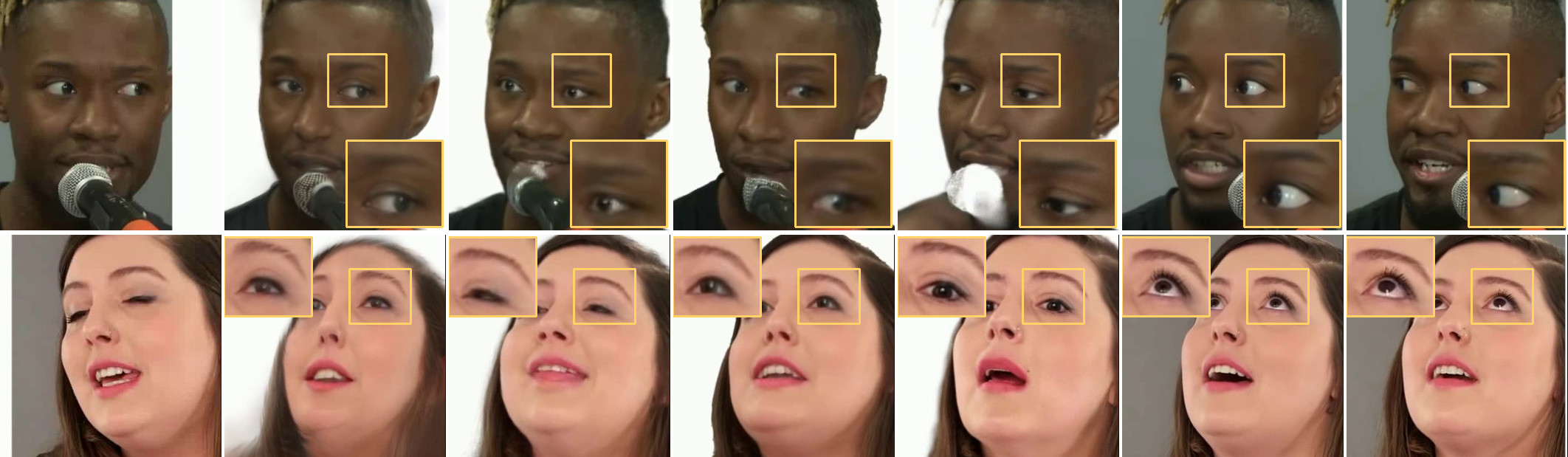}\\[2pt]
{\tiny
\makebox[0.1428\linewidth]{Ref.}%
\makebox[0.1428\linewidth]{LAM}%
\makebox[0.1428\linewidth]{FlexAvatar}%
\makebox[0.1428\linewidth]{UIKA}%
\makebox[0.1428\linewidth]{CAP4D}%
\makebox[0.1428\linewidth]{Ours}%
\makebox[0.1428\linewidth]{GT}}
\caption{Qualitative comparisons on the in-the-wild VFHQ test
set~\cite{xie2022vfhq}, against SotA 3D avatars.}
\label{fig:selfreenact-vfhq}
\end{figure}

\fi

\begin{figure*}[t]
\centering
\includegraphics[width=\linewidth]{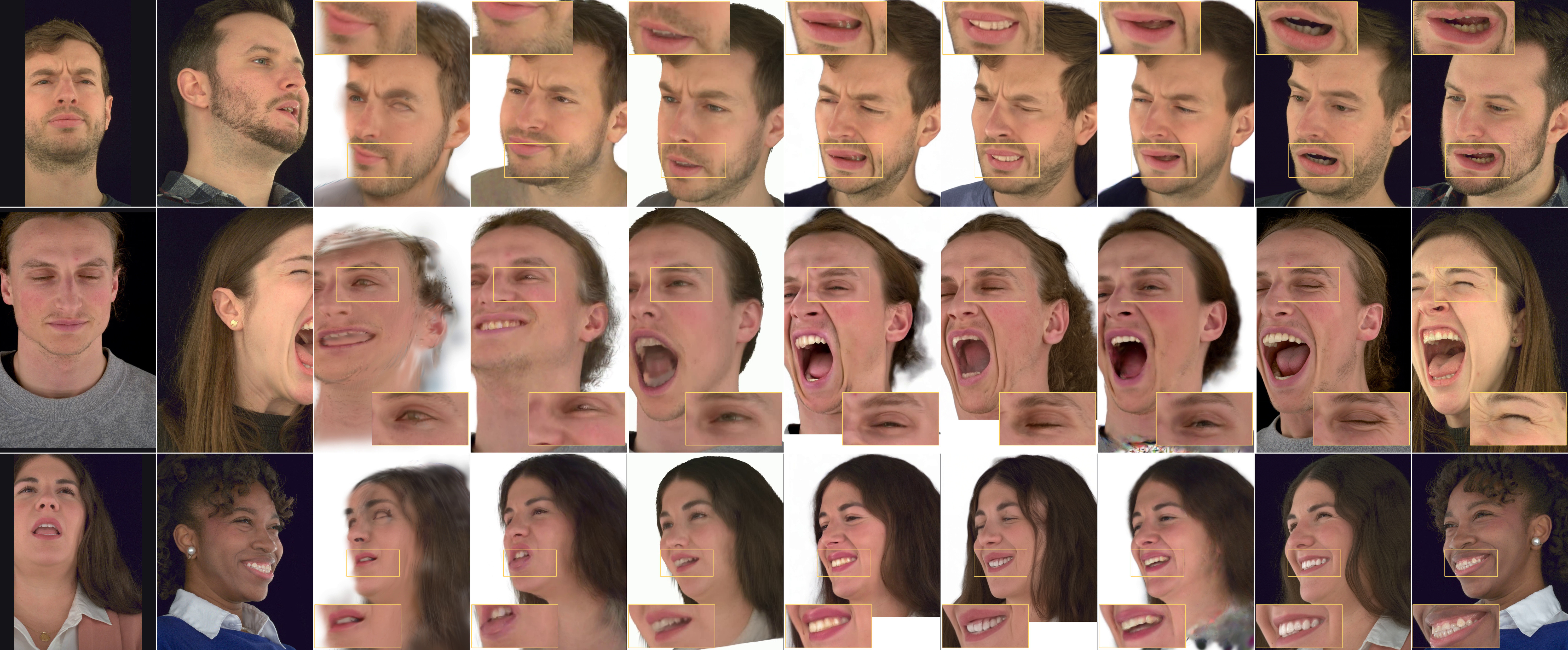}\\[1pt]
\vspace{-4pt}
\makebox[\linewidth][l]{\tiny
\makebox[0.09960\linewidth]{Reference}%
\hspace{0.00044\linewidth}%
\makebox[0.09960\linewidth]{Driving}%
\hspace{0.00044\linewidth}%
\makebox[0.09960\linewidth]{LAM}%
\hspace{0.00044\linewidth}%
\makebox[0.09960\linewidth]{FlexAvatar}%
\hspace{0.00044\linewidth}%
\makebox[0.09960\linewidth]{UIKA}%
\hspace{0.00044\linewidth}%
\makebox[0.09960\linewidth]{MMDM}%
\hspace{0.00044\linewidth}%
\makebox[0.09960\linewidth]{MVP4D}%
\hspace{0.00044\linewidth}%
\makebox[0.09960\linewidth]{CAP4D}%
\hspace{0.00044\linewidth}%
\makebox[0.09960\linewidth]{Ours}%
\hspace{0.00044\linewidth}%
\makebox[0.09960\linewidth]{Driver (target)}%
}
\vspace{-14pt}
\caption{Cross-reenactment on BecomingLit. The reference identity is driven
by a different person's expression and rendered to the target camera; the
last column shows the driver from that camera.}
\label{fig:crossreenact}
\end{figure*}

\ificlr\cparagraph{Cross-reenactment.}\else\subsection{Cross-Reenactment Results}\fi
Results appear in \cref{fig:crossreenact} and the rightmost
block of \cref{tab:selfreenact}. Without a ground-truth target, pixel
metrics are not applicable. Thus we measure identity similarity to the reference and expression and keypoint distance to the driver as seen from the target camera.

\ificlr
\begin{figure*}[t]
\centering
\includegraphics[width=\linewidth]{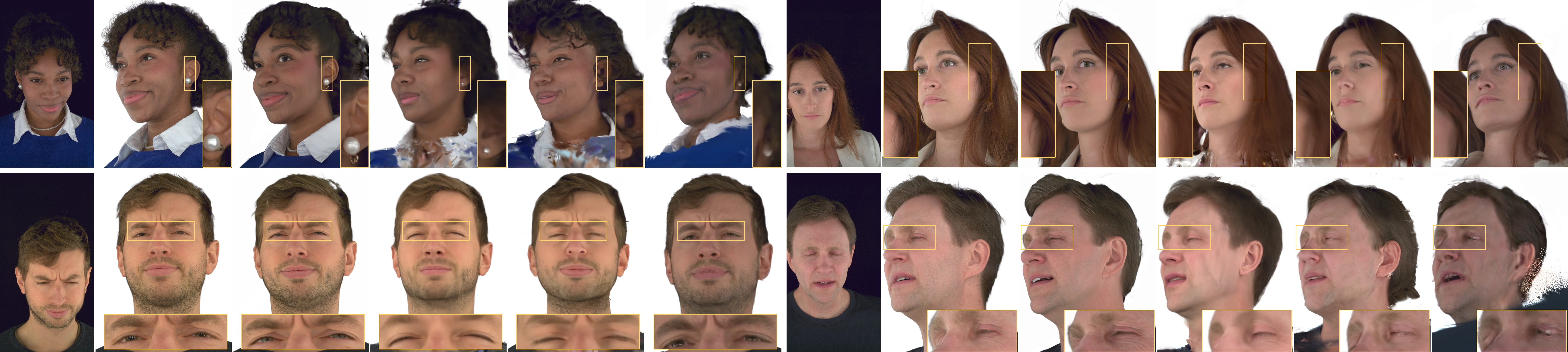}\\[1pt]
\vspace{-4pt}
\makebox[\linewidth][l]{\tiny
\makebox[0.06010\linewidth]{Reference}%
\hspace{0.00081\linewidth}%
\makebox[0.08683\linewidth]{GT}%
\hspace{0.00081\linewidth}%
\makebox[0.08683\linewidth]{Ours}%
\hspace{0.00081\linewidth}%
\makebox[0.08683\linewidth]{MMDM}%
\hspace{0.00081\linewidth}%
\makebox[0.08683\linewidth]{MVP4D}%
\hspace{0.00081\linewidth}%
\makebox[0.08683\linewidth]{FaceCam}%
\hspace{0.00325\linewidth}%
\makebox[0.06010\linewidth]{Reference}%
\hspace{0.00081\linewidth}%
\makebox[0.08683\linewidth]{GT}%
\hspace{0.00081\linewidth}%
\makebox[0.08683\linewidth]{Ours}%
\hspace{0.00081\linewidth}%
\makebox[0.08683\linewidth]{MMDM}%
\hspace{0.00081\linewidth}%
\makebox[0.08683\linewidth]{MVP4D}%
\hspace{0.00081\linewidth}%
\makebox[0.08683\linewidth]{FaceCam}%
}
\vspace{-14pt}
\caption{Multi-view consistency probe: each method's generation along a novel-view
ellipse is fit into a static 3DGS and re-rendered at a rig camera (vs.\ GT).
Quantitative results in \cref{tab:traj3dgs}.}
\label{fig:traj3dgs}
\end{figure*}

\fi

\ificlr\cparagraph{Multi-view consistency probe.}\else\subsection{Novel-View and Multi-View Consistency Results}\fi
\label{sec:novelview}

\ificlr
\begin{figure*}[t]
\centering
\includegraphics[width=\linewidth]{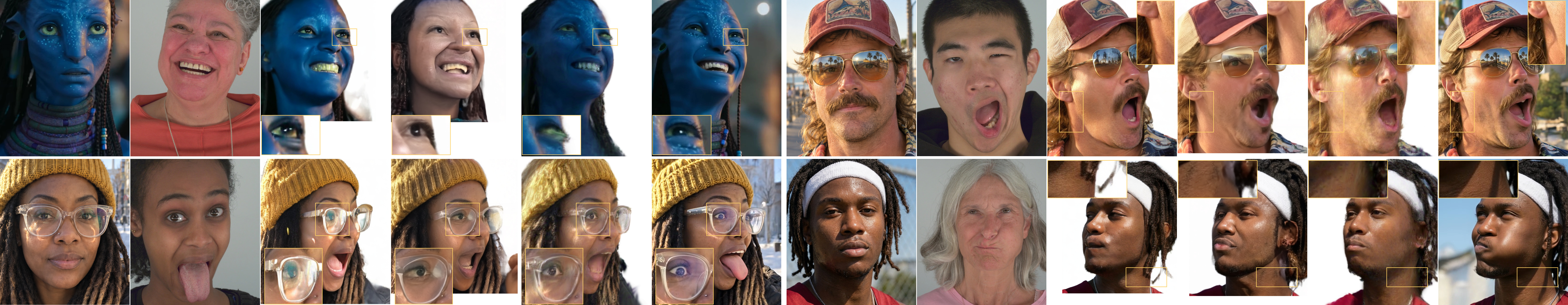}\\[1pt]
\vspace{-4pt}
\makebox[\linewidth][l]{\tiny
\makebox[0.08267\linewidth]{Reference}%
\hspace{0.00046\linewidth}%
\makebox[0.08267\linewidth]{Driving}%
\hspace{0.00046\linewidth}%
\makebox[0.08267\linewidth]{MMDM}%
\hspace{0.00046\linewidth}%
\makebox[0.08267\linewidth]{MVP4D}%
\hspace{0.00046\linewidth}%
\makebox[0.08267\linewidth]{FlexAvatar}%
\hspace{0.00046\linewidth}%
\makebox[0.08267\linewidth]{Ours}%
\hspace{0.00326\linewidth}%
\makebox[0.08267\linewidth]{Reference}%
\hspace{0.00046\linewidth}%
\makebox[0.08267\linewidth]{Driving}%
\hspace{0.00046\linewidth}%
\makebox[0.08267\linewidth]{MMDM}%
\hspace{0.00046\linewidth}%
\makebox[0.08267\linewidth]{MVP4D}%
\hspace{0.00046\linewidth}%
\makebox[0.08267\linewidth]{FlexAvatar}%
\hspace{0.00046\linewidth}%
\makebox[0.08267\linewidth]{Ours}%
}
\vspace{-14pt}
\caption{In-the-wild novel-view reenactment: the camera sweeps a wide arc,
isolating viewpoint consistency from expression transfer.}
\label{fig:trajbench}
\vspace{-10pt}
\end{figure*}

\begin{figure*}[t]
\centering
{\tiny
\def\spinL{7pt}%
\edef\spinu{\the\dimexpr(\linewidth-2\dimexpr\spinL\relax-1pt)/9274\relax}%
\edef\spinRef{\the\dimexpr\spinu*1125\relax}%
\edef\spinGrid{\the\dimexpr\spinu*3500\relax}%
\edef\spinH{\the\dimexpr\spinu*2000\relax}%
\noindent\makebox[\linewidth][c]{%
\parbox[t]{\spinRef}{\includegraphics[width=\spinRef,height=\spinH]{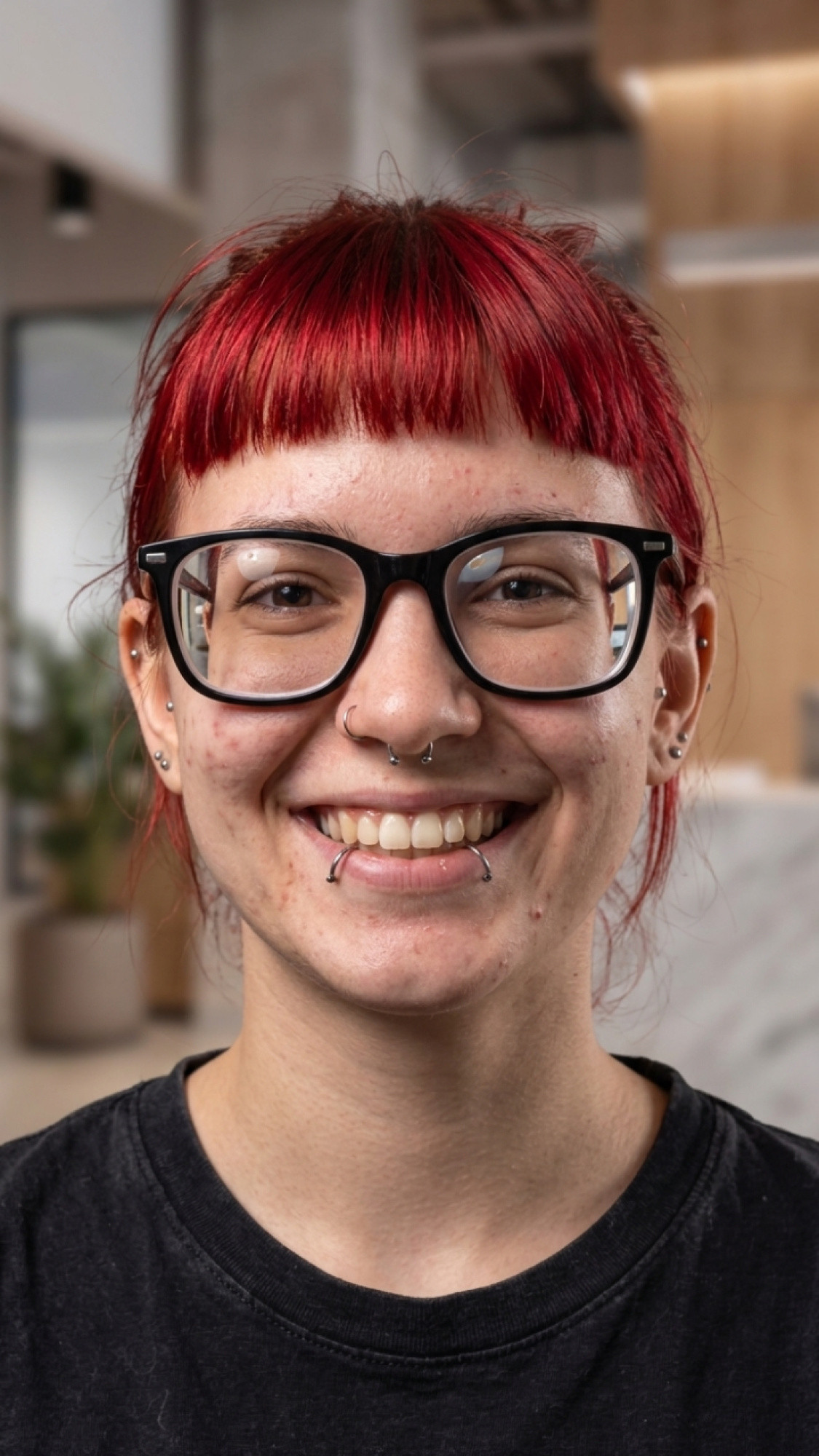}\\[1pt]\makebox[\spinRef]{Reference}}%
\makebox[\spinL]{\hfil\rotatebox{90}{\makebox[\spinH]{\makebox[\dimexpr\spinH/4\relax]{\adjustbox{max width=\dimexpr\spinH/4-1pt\relax}{Ours}}\makebox[\dimexpr\spinH/4\relax]{\adjustbox{max width=\dimexpr\spinH/4-1pt\relax}{MVP4D}}\makebox[\dimexpr\spinH/4\relax]{\adjustbox{max width=\dimexpr\spinH/4-1pt\relax}{FaceCam}}\makebox[\dimexpr\spinH/4\relax]{\adjustbox{max width=\dimexpr\spinH/4-1pt\relax}{FlexAvatar}}}}\hfil}%
\parbox[t]{\spinGrid}{\includegraphics[width=\spinGrid,height=\spinH]{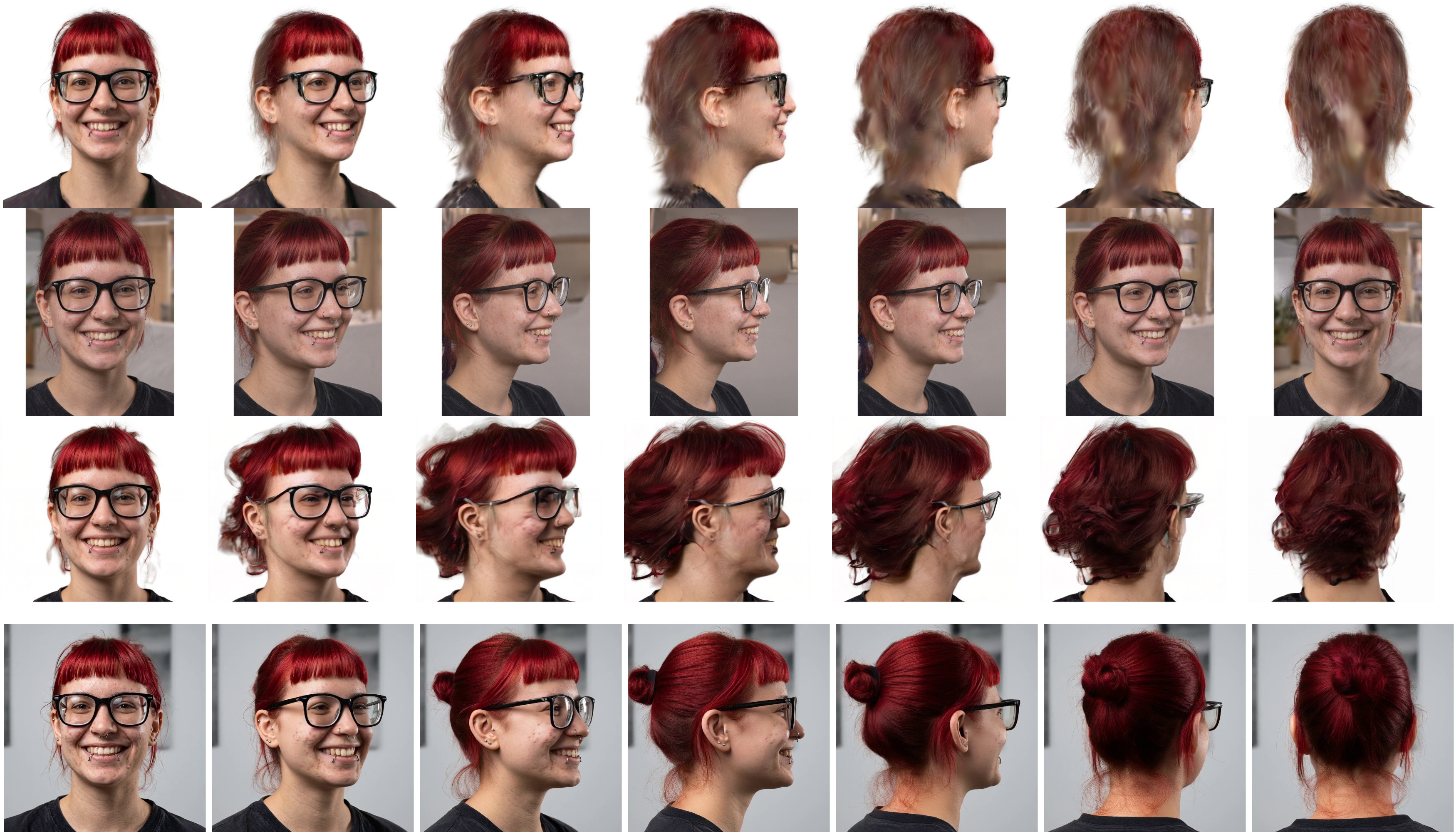}\\[1pt]\makebox[\dimexpr\spinGrid/7\relax]{$0^\circ$}\makebox[\dimexpr\spinGrid/7\relax]{$30^\circ$}\makebox[\dimexpr\spinGrid/7\relax]{$60^\circ$}\makebox[\dimexpr\spinGrid/7\relax]{$90^\circ$}\makebox[\dimexpr\spinGrid/7\relax]{$120^\circ$}\makebox[\dimexpr\spinGrid/7\relax]{$150^\circ$}\makebox[\dimexpr\spinGrid/7\relax]{$180^\circ$}}%
\hspace{\dimexpr\spinu*24\relax}%
\parbox[t]{\spinRef}{\includegraphics[width=\spinRef,height=\spinH]{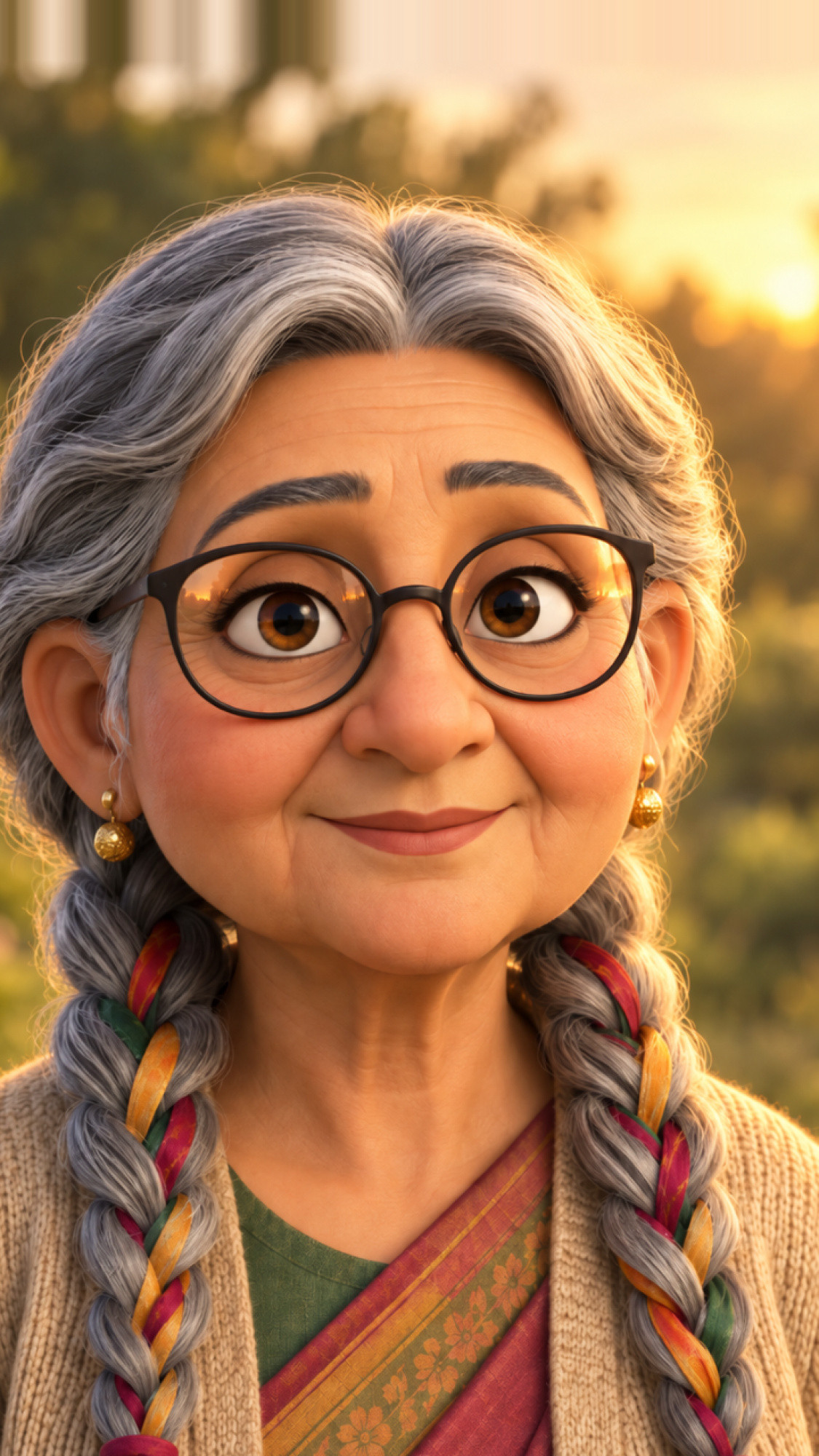}\\[1pt]\makebox[\spinRef]{Reference}}%
\makebox[\spinL]{\hfil\rotatebox{90}{\makebox[\spinH]{\makebox[\dimexpr\spinH/4\relax]{\adjustbox{max width=\dimexpr\spinH/4-1pt\relax}{Ours}}\makebox[\dimexpr\spinH/4\relax]{\adjustbox{max width=\dimexpr\spinH/4-1pt\relax}{MVP4D}}\makebox[\dimexpr\spinH/4\relax]{\adjustbox{max width=\dimexpr\spinH/4-1pt\relax}{FaceCam}}\makebox[\dimexpr\spinH/4\relax]{\adjustbox{max width=\dimexpr\spinH/4-1pt\relax}{FlexAvatar}}}}\hfil}%
\parbox[t]{\spinGrid}{\includegraphics[width=\spinGrid,height=\spinH]{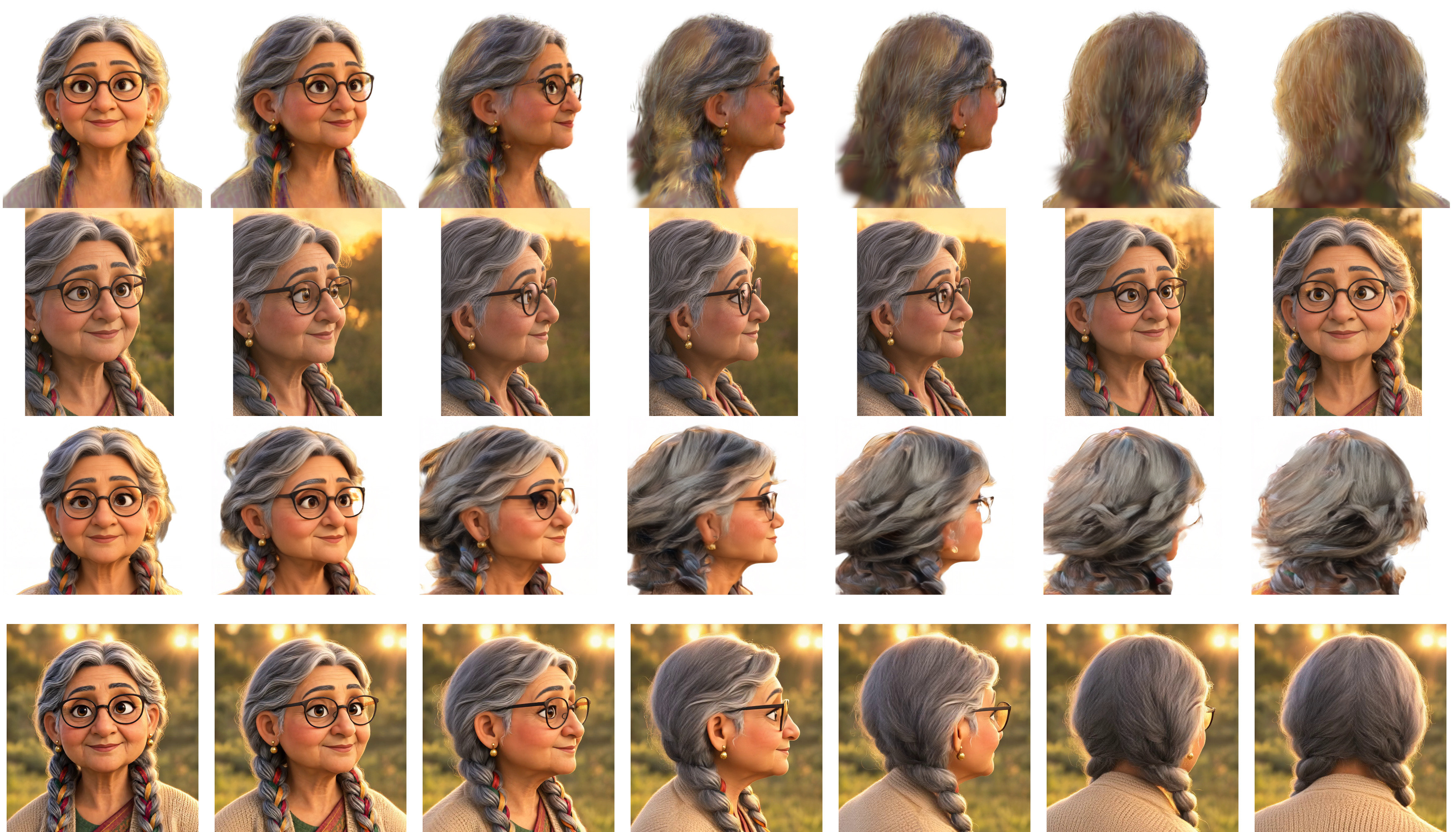}\\[1pt]\makebox[\dimexpr\spinGrid/7\relax]{$0^\circ$}\makebox[\dimexpr\spinGrid/7\relax]{$30^\circ$}\makebox[\dimexpr\spinGrid/7\relax]{$60^\circ$}\makebox[\dimexpr\spinGrid/7\relax]{$90^\circ$}\makebox[\dimexpr\spinGrid/7\relax]{$120^\circ$}\makebox[\dimexpr\spinGrid/7\relax]{$150^\circ$}\makebox[\dimexpr\spinGrid/7\relax]{$180^\circ$}}%
}}
\vspace{-4pt}
\caption{360\textdegree{} novel-view synthesis from a single reference
image: the camera orbits from the front ($0^\circ$) to the back of the head
($180^\circ$). Rows: FlexAvatar\tabcite{kirschstein2026flexavatar},
FaceCam\tabcite{facecam}, MVP4D\tabcite{mvp4d}, and ours.}
\label{fig:spin360}
\end{figure*}

\fi

The 3DGS probe (\cref{fig:traj3dgs}, \cref{tab:traj3dgs}) measures multi-view
consistency quantitatively: since the rig cameras lie off the generated
trajectory, a fit only scores well if the generations agree across views
and stay faithful to the subject. Our fits outperform every baseline on
every metric and are significantly sharper than all baseline fits, whose
view-inconsistent generations blur the reconstruction
(\cref{fig:traj3dgs}), confirming the multi-view consistency of \OURS.

\ificlr\cparagraph{Generalization.}\else\fi
\Cref{fig:trajbench} combines cross-reenactment with a wide camera sweep on
challenging in-the-wild references: cast shadows, reflections and
refractions in glasses, and human-like characters far outside every
method's training data. \Cref{fig:spin360} orbits the camera to the back of
the head, beyond the viewpoints of our training captures. 
Together, these results show that \OURS generalize to rendering effects,
subjects, and camera poses absent from the training data. This capacity
stems from the internet-scale video prior, which our conditioning is
designed to preserve. The \vdasuppvideo{} further shows clothing
dynamics, dangling earrings, and view-dependent lighting that follow the
head, effects that explicit 3D renderers struggle to reproduce.

\ificlr\else
\begin{figure*}[t]
\centering
\includegraphics[width=\linewidth]{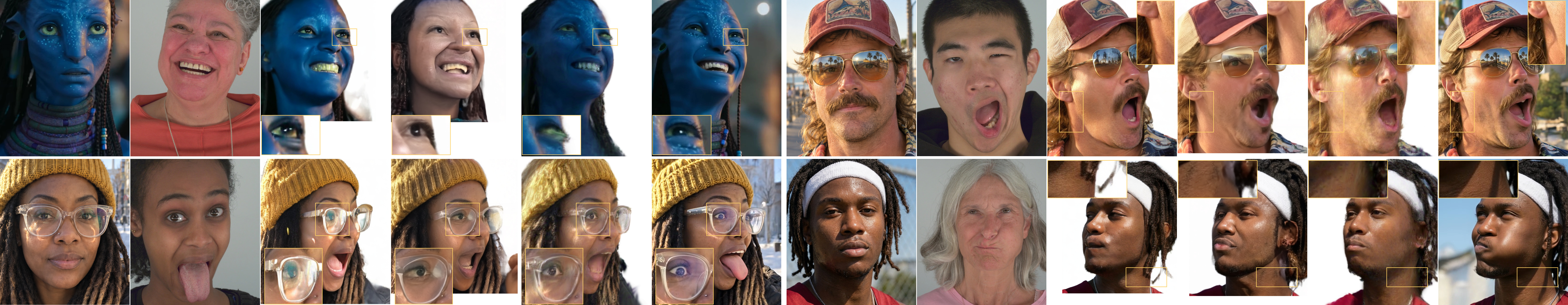}\\[1pt]
\vspace{-4pt}
\makebox[\linewidth][l]{\tiny
\makebox[0.08267\linewidth]{Reference}%
\hspace{0.00046\linewidth}%
\makebox[0.08267\linewidth]{Driving}%
\hspace{0.00046\linewidth}%
\makebox[0.08267\linewidth]{MMDM}%
\hspace{0.00046\linewidth}%
\makebox[0.08267\linewidth]{MVP4D}%
\hspace{0.00046\linewidth}%
\makebox[0.08267\linewidth]{FlexAvatar}%
\hspace{0.00046\linewidth}%
\makebox[0.08267\linewidth]{Ours}%
\hspace{0.00326\linewidth}%
\makebox[0.08267\linewidth]{Reference}%
\hspace{0.00046\linewidth}%
\makebox[0.08267\linewidth]{Driving}%
\hspace{0.00046\linewidth}%
\makebox[0.08267\linewidth]{MMDM}%
\hspace{0.00046\linewidth}%
\makebox[0.08267\linewidth]{MVP4D}%
\hspace{0.00046\linewidth}%
\makebox[0.08267\linewidth]{FlexAvatar}%
\hspace{0.00046\linewidth}%
\makebox[0.08267\linewidth]{Ours}%
}
\vspace{-14pt}
\caption{In-the-wild novel-view reenactment: the camera sweeps a wide arc,
isolating viewpoint consistency from expression transfer.}
\label{fig:trajbench}
\vspace{-10pt}
\end{figure*}

\fi

\ificlr\else
\begin{figure*}[t]
\centering
\includegraphics[width=\linewidth]{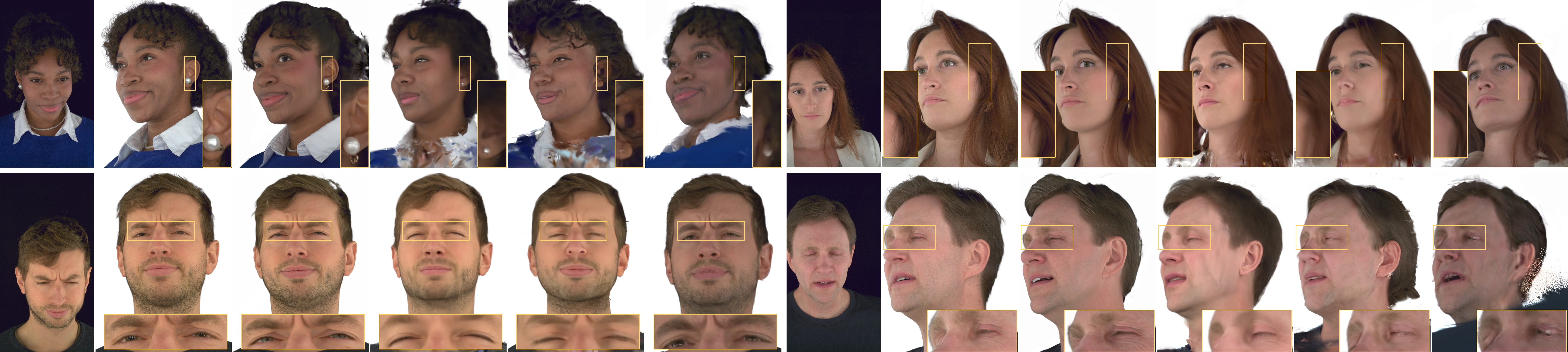}\\[1pt]
\vspace{-4pt}
\makebox[\linewidth][l]{\tiny
\makebox[0.06010\linewidth]{Reference}%
\hspace{0.00081\linewidth}%
\makebox[0.08683\linewidth]{GT}%
\hspace{0.00081\linewidth}%
\makebox[0.08683\linewidth]{Ours}%
\hspace{0.00081\linewidth}%
\makebox[0.08683\linewidth]{MMDM}%
\hspace{0.00081\linewidth}%
\makebox[0.08683\linewidth]{MVP4D}%
\hspace{0.00081\linewidth}%
\makebox[0.08683\linewidth]{FaceCam}%
\hspace{0.00325\linewidth}%
\makebox[0.06010\linewidth]{Reference}%
\hspace{0.00081\linewidth}%
\makebox[0.08683\linewidth]{GT}%
\hspace{0.00081\linewidth}%
\makebox[0.08683\linewidth]{Ours}%
\hspace{0.00081\linewidth}%
\makebox[0.08683\linewidth]{MMDM}%
\hspace{0.00081\linewidth}%
\makebox[0.08683\linewidth]{MVP4D}%
\hspace{0.00081\linewidth}%
\makebox[0.08683\linewidth]{FaceCam}%
}
\vspace{-14pt}
\caption{Multi-view consistency probe: each method's generation along a novel-view
ellipse is fit into a static 3DGS and re-rendered at a rig camera (vs.\ GT).
Quantitative results in \cref{tab:traj3dgs}.}
\label{fig:traj3dgs}
\end{figure*}

\fi

\ificlr\else
\begin{table}[t]
\centering
\caption{3DGS distillation ($n=13$): Evaluated against 16 G.T. cameras.}
\vspace{-5pt}
\label{tab:traj3dgs}
\footnotesize
\setlength{\tabcolsep}{1.5pt}
\begin{tabular}{lcccc}
\toprule
Method & PSNR~$\uparrow$ & LPIPS~$\downarrow$ & ID~$\uparrow$ & Expr~$\uparrow$ \\
\midrule
UIKA\tabcite{wu2026uika} & 21.96 & 0.244 & 0.697 & 0.868 \\
FlexAvatar\tabcite{kirschstein2026flexavatar} & \cellcolor{tabsecond}23.28 & \cellcolor{tabsecond}0.203 & \cellcolor{tabsecond}0.753 & \cellcolor{tabsecond}0.941 \\
MMDM\tabcite{taubner2025cap4d} & \cellcolor{tabsecond}23.28 & \cellcolor{tabthird}0.242 & 0.722 & 0.871 \\
MVP4D\tabcite{mvp4d} & \cellcolor{tabthird}21.99 & 0.275 & 0.687 & 0.889 \\
FaceCam\tabcite{facecam} & 18.86 & 0.292 & \cellcolor{tabthird}0.731 & \cellcolor{tabthird}0.922 \\
Ours & \cellcolor{tabfirst}23.69 & \cellcolor{tabfirst}0.192 & \cellcolor{tabfirst}0.869 & \cellcolor{tabfirst}0.968 \\
\bottomrule
\end{tabular}
\end{table}

\fi

\ificlr\else

\fi

\ificlr\cparagraph{Ablations.}\else\subsection{Ablations}\fi
\label{sec:ablations}

\begin{figure*}[t]
\centering
\includegraphics[width=\linewidth]{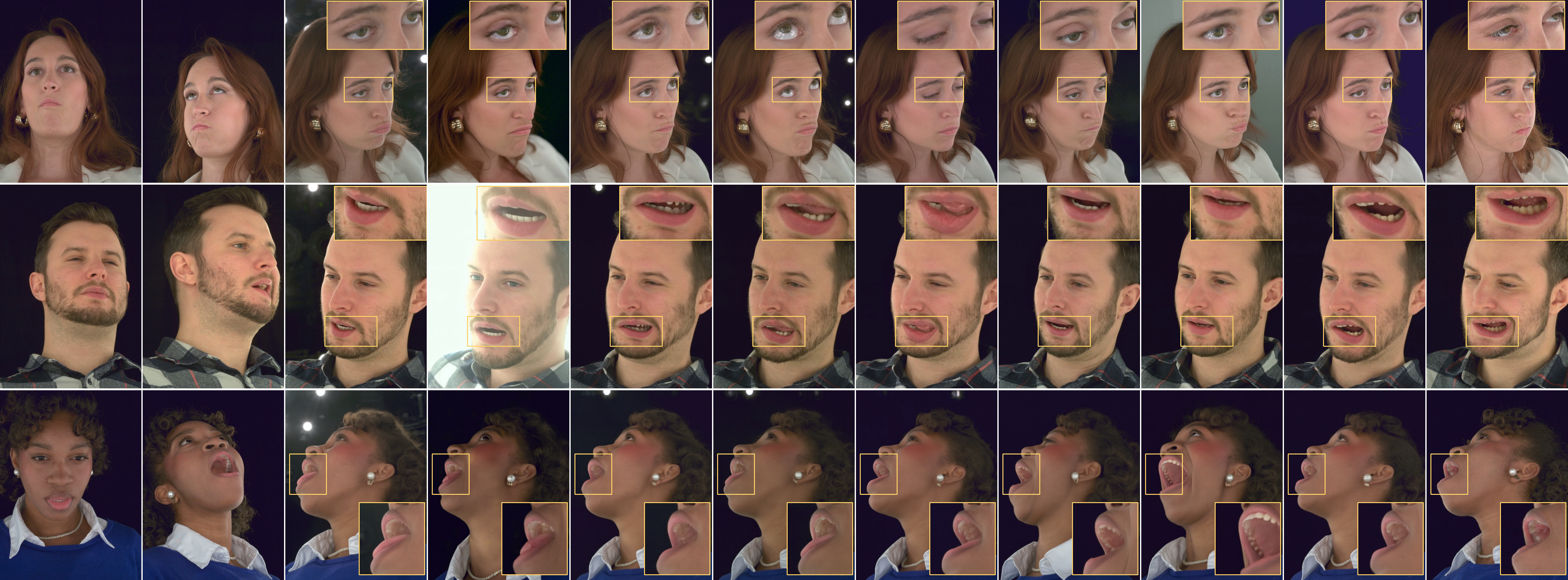}\\[1pt]
\vspace{-4pt}
\makebox[\linewidth][l]{\tiny
\makebox[0.09017\linewidth]{Reference}%
\hspace{0.00081\linewidth}%
\makebox[0.09017\linewidth]{Driving}%
\hspace{0.00081\linewidth}%
\makebox[0.09017\linewidth]{WAN-A.\ expr.}%
\hspace{0.00081\linewidth}%
\makebox[0.09017\linewidth]{$+$ fr.\ fusers}%
\hspace{0.00081\linewidth}%
\makebox[0.09017\linewidth]{No disent.}%
\hspace{0.00081\linewidth}%
\makebox[0.09017\linewidth]{Self-driven}%
\hspace{0.00081\linewidth}%
\makebox[0.09017\linewidth]{No expr.}%
\hspace{0.00081\linewidth}%
\makebox[0.09017\linewidth]{No mesh}%
\hspace{0.00081\linewidth}%
\makebox[0.09017\linewidth]{Rand.\ corpus}%
\hspace{0.00081\linewidth}%
\makebox[0.09017\linewidth]{Ours}%
\hspace{0.00081\linewidth}%
\makebox[0.09017\linewidth]{Ground truth}%
}
\vspace{-14pt}
\caption{Qualitative ablation on BecomingLit v3; columns match
\cref{tab:ablation-self}.}
\label{fig:ablation}
\vspace{-10pt}
\end{figure*}

\Cref{tab:ablation-self,fig:ablation} analyze our design choices on
BecomingLit self-reenactment. For more details on the exact settings, we refer to \cref{sec:supp-baselines}. Notably, the frozen WAN-Animate base model performs poorly at novel-view self-reenactment, and replacing its expression code $\zc{WAN}{raw}$ with the viewpoint-neutral code improves the metrics only slightly.\\
\begin{table*}[t]
\centering
\caption{\textbf{Self-reenactment ablations} on BecomingLit v3 ($n$=25), the only ablation dataset.}
\vspace{-4pt}
\label{tab:ablation-self}
\resizebox{\linewidth}{!}{%
\footnotesize
\setlength{\tabcolsep}{4pt}
\setlength{\fboxsep}{0.6pt}
\begin{tabular}{lccccccccccc}
\toprule
& \multicolumn{4}{c}{Ablated knob} & \multicolumn{7}{c}{Self-reenactment, BecomingLit v3 ($n$=25)} \\
\cmidrule(lr){2-5}\cmidrule(lr){6-12}
Method & Fine-tune & Expr.\ code & Pose cond. & Disent. & PSNR~$\uparrow$ & LPIPS~$\downarrow$ & JOD~$\uparrow$ & ID~$\uparrow$ & Expr.~$\uparrow$ & AKD~$\downarrow$ & APD~$\downarrow$ \\
\midrule
WAN-A. & none & WAN native & 2D skel. & -- & 16.87 & 0.2582 & 3.897 & 0.5825 & 0.8279 & 0.0813 & 6.06 \\
WAN-A. disent. & none & WAN native & 2D skel. & \checkmark & 17.03 & 0.2584 & 3.934 & 0.6044 & 0.8386 & 0.0791 & 5.72 \\
\midrule
Ours, WAN-A. expr. & all but enc. & WAN native & skel.\ $+$ mesh & -- & 21.52 & 0.2098 & 4.438 & \colorbox{tabfirst}{0.7012} & \colorbox{tabthird}{0.9115} & \colorbox{tabfirst}{0.0532} & 4.06 \\
\quad $+$ frozen fusers & all but enc., fusers & WAN native & skel.\ $+$ mesh & -- & 19.67 & 0.2806 & 4.052 & 0.6534 & 0.8175 & 0.0808 & 6.03 \\
\midrule
Ours, no disent.\ $\dagger$ & full & X-NeMo & skel.\ $+$ mesh & $\times$ & \colorbox{tabsecond}{22.06} & \colorbox{tabsecond}{0.2048} & 4.608 & \colorbox{tabthird}{0.6927} & 0.9099 & \colorbox{tabthird}{0.0550} & \colorbox{tabfirst}{3.94} \\
\quad self-driven $\dagger$ & full & X-NeMo & skel.\ $+$ mesh & $\times$ & 20.69 & 0.2207 & \colorbox{tabsecond}{4.613} & 0.6866 & 0.8744 & 0.0608 & 4.51 \\
\midrule
Ours, no expr.\ codes & full & zeroed & skel.\ $+$ mesh & -- & 21.61 & \colorbox{tabthird}{0.2052} & \colorbox{tabthird}{4.609} & 0.6896 & 0.8796 & 0.0561 & \colorbox{tabsecond}{3.96} \\
Ours, no proxy mesh & full & X-NeMo & 2D skel. & \checkmark & 18.34 & 0.2503 & 4.196 & 0.6840 & \colorbox{tabsecond}{0.9124} & 0.0722 & 4.93 \\
Ours, rand. corpus & full & X-NeMo & skel.\ $+$ mesh & \checkmark & \colorbox{tabthird}{21.71} & 0.2487 & 4.462 & 0.6881 & 0.8954 & 0.0661 & 4.75 \\
\midrule
Ours (full) & full & X-NeMo & skel.\ $+$ mesh & \checkmark & \colorbox{tabfirst}{22.43} & \colorbox{tabfirst}{0.2021} & \colorbox{tabfirst}{4.638} & \colorbox{tabsecond}{0.6989} & \colorbox{tabfirst}{0.9201} & \colorbox{tabsecond}{0.0549} & \colorbox{tabthird}{3.97} \\
\bottomrule
\end{tabular}%
}
\end{table*}

\emph{Expression codes.} Our X-NeMo codes outperform WAN-Animate's native
expression code in image quality and expression fidelity, validating our
choice. Nevertheless, the native-code variant remains competitive, showing
that our recipe works with different latent expression spaces and can adopt
stronger encoders in the future. Without any expression code, the model
misses subtle expressions (\cref{fig:ablation}), which are not captured by FLAME.
\emph{Data and pose.} Training on a random instead of our curated corpus
degrades every metric, and conditioning only on a ViT-Pose skeleton, WAN-Animate's
native pose signal, without the proxy mesh costs 4.1\,dB PSNR and raises AKD from 0.055 to 0.072, while
expression fidelity stays high (0.912 vs.\ 0.920).
\emph{Disentanglement.} Without disentangler and camera triplets
(self-driven: $\pi_{\text{drv}}=\pi_{\text{tar}}$), performance drops
markedly. Without the disentangler but with
camera triplets, the model is almost on par with our full approach.
Remarkably, triplet supervision alone enforces view-invariant expression
codes, raising DISTRACT rank-1 accuracy from 0.244 to 0.792 during
training, whereas self-driven training stalls at 0.486
(\cref{fig:disent-emergence}, \cref{sec:supp-emergence}).

\ificlr\cparagraph{FactorPortrait comparison.}\else\subsection{Comparison to FactorPortrait}\fi
\label{sec:factorportrait}

\ificlr
\begin{table}[t]
\centering
\caption{FactorPortrait comparison (face-region metrics, \cref{sec:metrics}).}
\vspace{-4pt}
\label{tab:factorportrait}
\footnotesize
\setlength{\tabcolsep}{2.6pt}
\setlength{\fboxsep}{0.6pt}
\begin{adjustbox}{max width=\linewidth}
\begin{tabular}{lcccccccc}
\toprule
Method & PSNR~$\uparrow$ & SSIM~$\uparrow$ & LPIPS~$\downarrow$ & JOD~$\uparrow$ & AKD~$\downarrow$ & APD~$\downarrow$ & ID~$\uparrow$ & Expr~$\uparrow$ \\
\midrule
FactorPortrait~\cite{tang2025factorportrait} & 24.71 & 0.712 & 0.0208 & 7.02 & 0.0329 & \colorbox{tabfirst}{2.37} & 0.747 & 0.946 \\
Ours & \colorbox{tabfirst}{24.75} & \colorbox{tabfirst}{0.717} & \colorbox{tabfirst}{0.0190} & \colorbox{tabfirst}{7.14} & \colorbox{tabfirst}{0.0322} & 2.94 & \colorbox{tabfirst}{0.803} & \colorbox{tabfirst}{0.969} \\
\bottomrule
\end{tabular}
\end{adjustbox}
\vspace{-10pt}
\end{table}

\begin{figure*}[t]
\centering
\includegraphics[width=\linewidth]{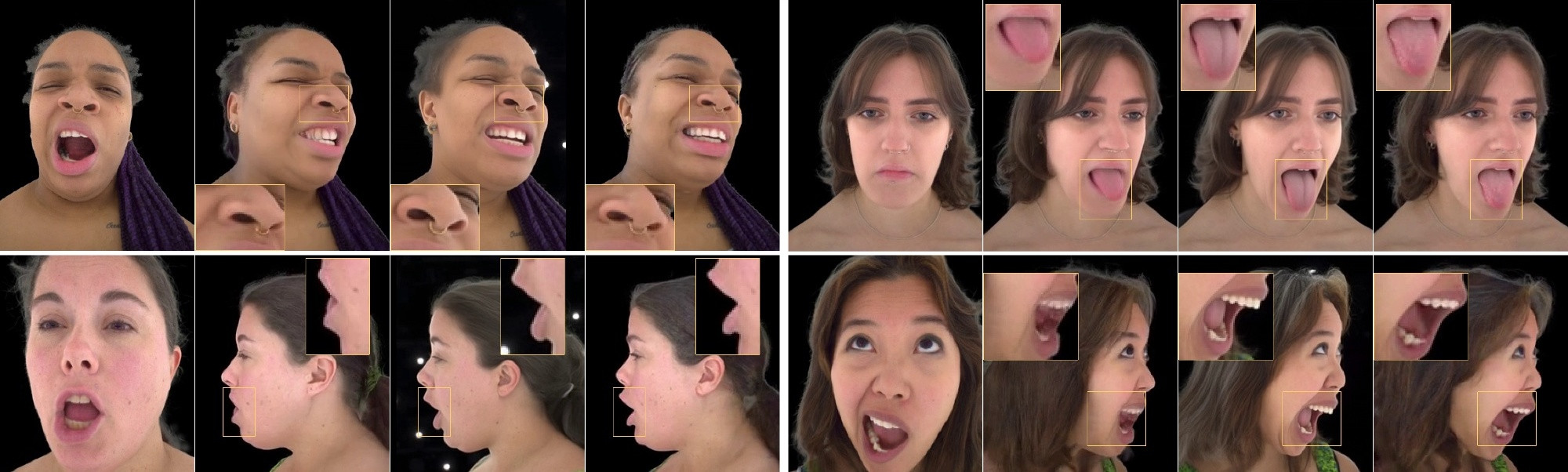}\\[1pt]
{\footnotesize
\makebox[0.4975\linewidth]{%
\makebox[0.1244\linewidth]{Reference}%
\makebox[0.1244\linewidth]{FactorPortrait}%
\makebox[0.1244\linewidth]{Ours}%
\makebox[0.1244\linewidth]{Ground truth}}\hfill
\makebox[0.4975\linewidth]{%
\makebox[0.1244\linewidth]{Reference}%
\makebox[0.1244\linewidth]{FactorPortrait}%
\makebox[0.1244\linewidth]{Ours}%
\makebox[0.1244\linewidth]{Ground truth}}}
\vspace{-14pt}
\caption{Comparison to FactorPortrait on its released examples.}
\label{fig:factorportrait}
\end{figure*}

\fi

FactorPortrait~\cite{tang2025factorportrait} is concurrent work that also
controls expression, pose, and viewpoint in a VDM. No code or model is
available and the authors confirmed that none can be shared, so we run our
method on the self-reenactment examples released with their paper. These examples follow FactorPortrait's
rather specific data distribution, which favors FactorPortrait. Nevertheless, \OURS slightly improve
image quality and clearly improve identity preservation and expression
fidelity (\cref{tab:factorportrait}). \Cref{fig:factorportrait} further
shows that FactorPortrait degrades under strong viewpoint changes.

\ificlr\else
\begin{table}[t]
\centering
\caption{FactorPortrait comparison (face-region metrics, \cref{sec:metrics}).}
\vspace{-4pt}
\label{tab:factorportrait}
\footnotesize
\setlength{\tabcolsep}{2.6pt}
\setlength{\fboxsep}{0.6pt}
\begin{adjustbox}{max width=\linewidth}
\begin{tabular}{lcccccccc}
\toprule
Method & PSNR~$\uparrow$ & SSIM~$\uparrow$ & LPIPS~$\downarrow$ & JOD~$\uparrow$ & AKD~$\downarrow$ & APD~$\downarrow$ & ID~$\uparrow$ & Expr~$\uparrow$ \\
\midrule
FactorPortrait~\cite{tang2025factorportrait} & 24.71 & 0.712 & 0.0208 & 7.02 & 0.0329 & \colorbox{tabfirst}{2.37} & 0.747 & 0.946 \\
Ours & \colorbox{tabfirst}{24.75} & \colorbox{tabfirst}{0.717} & \colorbox{tabfirst}{0.0190} & \colorbox{tabfirst}{7.14} & \colorbox{tabfirst}{0.0322} & 2.94 & \colorbox{tabfirst}{0.803} & \colorbox{tabfirst}{0.969} \\
\bottomrule
\end{tabular}
\end{adjustbox}
\vspace{-10pt}
\end{table}

\begin{figure*}[t]
\centering
\includegraphics[width=\linewidth]{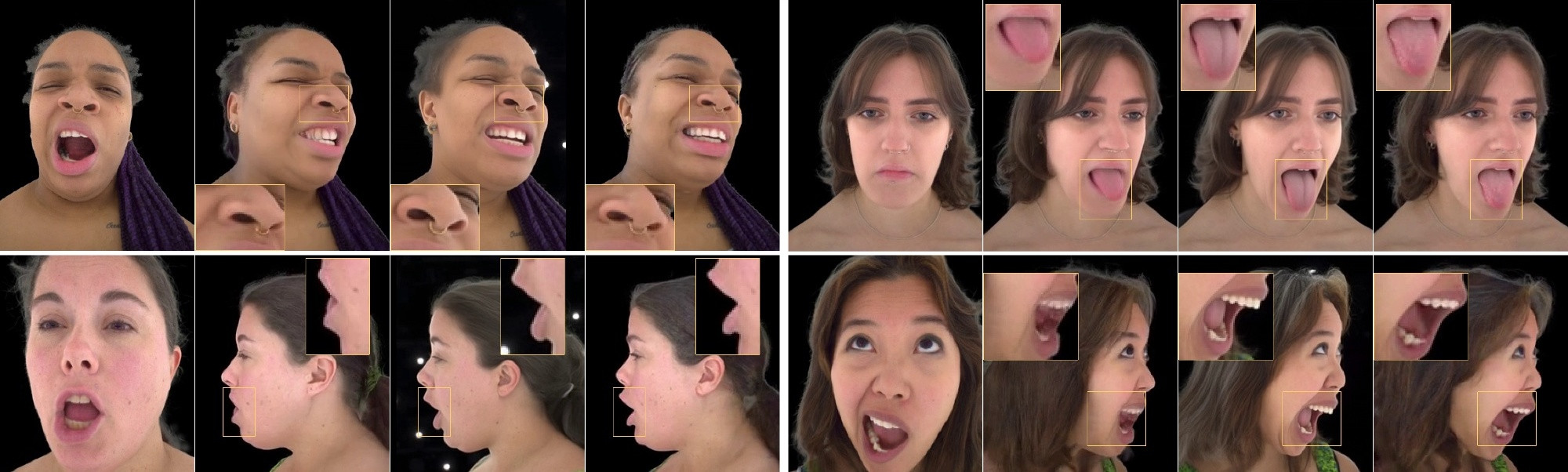}\\[1pt]
{\footnotesize
\makebox[0.4975\linewidth]{%
\makebox[0.1244\linewidth]{Reference}%
\makebox[0.1244\linewidth]{FactorPortrait}%
\makebox[0.1244\linewidth]{Ours}%
\makebox[0.1244\linewidth]{Ground truth}}\hfill
\makebox[0.4975\linewidth]{%
\makebox[0.1244\linewidth]{Reference}%
\makebox[0.1244\linewidth]{FactorPortrait}%
\makebox[0.1244\linewidth]{Ours}%
\makebox[0.1244\linewidth]{Ground truth}}}
\vspace{-14pt}
\caption{Comparison to FactorPortrait on its released examples.}
\label{fig:factorportrait}
\end{figure*}

\fi

\section{Limitations and Future Work}
\label{sec:limitations}

Unlike a 3DGS avatar, which renders in real time once built, \OURS pay
the cost of VDM inference for every clip (about 12 minutes for 77 frames at
$576\times1024$ on one A100), and long sequences are prone to temporal drift.
Autoregressive video and world
models~\cite{liu2026rollingforcing,samuel2026fastar,zhu2026causalforcing,zhao2026causalforcingpp},
including a real-time variant of WAN-Animate-2~\cite{cheng2025wananimate2},
have shown that long, consistent generation can be made fast. Motivated by
this, we distill our model with Self Forcing~\cite{huang2025selfforcing}
into a four-step auto-regressive VDM for
temporal consistency as a prototype (\cref{sec:supp-streaming}). This unoptimized implementation generates
4.8 frames per second at $288\times512$ on one A100 and stays temporally
stable over more than a thousand frames, at the cost of a small loss in
quality. Furthermore, lighting is baked into the reference image, and we rely on a fixed expression space rather than training it end-to-end. Our recipe is anchored to WAN-Animate-14B, although we expect it to transfer to
other backbones, as well.

\section{Conclusion}
\label{sec:conclusion}

We showed that an avatar does not have to be a 3D asset: by making a video
diffusion model controllable in expression, pose, and camera viewpoint,
\OURS turn its internet-scale prior into a single-image avatar that is
multi-view consistent and more faithful than both 3DGS-based and
diffusion-based avatars. Central to this is expression control through
learned latent spaces: our recipe works with different expression spaces,
and those of image encoders capture expressions with higher fidelity than
3DMMs, but must first be disentangled from camera viewpoint for the best
results. Training is affordable and accessible, taking 34 hours on a single
GPU with public multi-view data only. Because the prior is preserved rather
than distilled into 3D, \OURS render what explicit renderers struggle with
(shadows, reflections, refractions, hair dynamics, and clothing that
deforms with motion) and generalize to subjects, effects, and viewpoints
absent from the training data. We see this as evidence that controllable
video models are a practical foundation for photorealistic avatars.

\providecommand{\FILLIN}[1]{\textbf{\color{red}[#1]}}

\section*{Ethics Statement}

\OURS turn a single photograph into a photo-realistic, controllable video of
the depicted person. The same capability enables impersonation and
non-consensual synthetic media. We train and evaluate only on multi-view
datasets whose participants gave informed consent and which their authors
released for research use
(NeRSemble~\cite{kirschstein2023nersemble},
AVA-256~\cite{martinez2024codecavatarstudio},
BecomingLit~\cite{schmidt2025becominglit},
Multiface~\cite{wuu2022multiface}) and on the public VFHQ
benchmark~\cite{xie2022vfhq}.

\section*{Reproducibility Statement}

We will release our full code, covering data preprocessing, training,
inference, and evaluation, together with the trained model checkpoints\ifdefined\vdaarxiv{} at
\url{\vdacodeurl}\fi. All
training data (NeRSemble~\cite{kirschstein2023nersemble},
AVA-256~\cite{martinez2024codecavatarstudio}) and evaluation data
(BecomingLit~\cite{schmidt2025becominglit}, Multiface~\cite{wuu2022multiface},
VFHQ~\cite{xie2022vfhq}) are public research datasets. \Cref{sec:method}
describes the architecture and training objective, \cref{sec:supp-impl} lists
the training and inference hyperparameters, \cref{sec:supp-proxymesh} details
the proxy-mesh construction, and \cref{sec:supp-disent,sec:disentangle_training}
the disentangler. \Cref{sec:supp-evalprotocol,sec:metrics} define the
evaluation protocol, including the DISTRACT benchmark, and every metric;
\cref{sec:supp-baselines} describes how each baseline and ablation was run.

\section*{AI Use Statement}

In this work, we used generative AI tools to implement methods, as coding
agents that followed detailed, author-written specifications, and to generate
synthetic data, namely a small set of AI-generated portrait images that serve
as reference inputs for qualitative examples; these images are used neither
for training nor for any quantitative evaluation, but merely to demonstrate interesting and challenging inputs. We have not used generative AI tools to develop theoretical models or
conceptual frameworks, to formulate or prove mathematical claims, to assist
with translation, or to clean and reformat datasets, and the rest of the
required-disclosure tasks are not applicable to this work. Additionally, we
used generative AI tools to create and edit software code, including the
scripts that build the figures and tables (under careful supervision), to draft and edit parts of the
paper, and to search the literature. For the writing, the authors wrote
detailed bullet-point outlines, used generative AI to turn them into prose,
and then revised the text manually over several iterations. We have reviewed
all AI-assisted work: the authors reviewed and tested the generated code, our
own evaluation pipeline produces every number in the paper and the authors
checked each number against its source files, and the authors read and
revised all AI-assisted text. We take responsibility for the final content of
this work, including text, claims, or artifacts produced with the aid of
generative AI.

\ifdefined\vdaarxiv\section*{Acknowledgments}

This work was supported by Toyota Motor Europe and Woven by Toyota.
This work was also supported by the ERC Consolidator Grant Gen3D (101171131).
\fi

\FloatBarrier
\bibliographystyle{iclr2027_conference}
\bibliography{main}

\appendix

\makeatletter\let\iclr@appsection\section
\renewcommand{\section}{\FloatBarrier\iclr@appsection}\makeatother

\section{Evaluation Metrics}
\label{sec:metrics}

\section{Evaluation Protocol Details}
\label{sec:supp-evalprotocol}

\paragraph{DISTRACT.}
DISTRACT is built on the FLAME-tracked sequences of
NeRSemble~\cite{kirschstein2023nersemble} (all 16 cameras, sentence-reading
sequences excluded). Each query has one positive and four distractors. In the
\emph{viewpoint-invariance} task (500 queries), query and positive are two
cameras of the same identity, sequence and timestep. In the
\emph{identity-invariance} task, the positive is the nearest frame of another
identity in FLAME expression and jaw parameters; of 456 mined sets, we kept the
126 whose positive we manually verified to show the query's expression.
Distractors are random frames that share with the query its identity and camera
(A), only its identity (B), only its camera (C), or neither (D). Codes are
compared by cosine distance; Rank-1 is the fraction of queries for which no
distractor is closer than the positive, and Err-X the fraction for which
distractor X is. The MLP probe (two layers, supervised contrastive
loss~\cite{khosla2020supcon}, all views of one frame form a class) is fitted on
each encoder's NeRSemble embeddings, which include the benchmark frames; it is
thus transductive and measures how much view-invariant expression information
the frozen latents contain, not generalization to unseen subjects.

\paragraph{Reenactment evaluation.}
BecomingLit~\cite{schmidt2025becominglit} and Multiface~\cite{wuu2022multiface}
are part of neither our training data nor that of any compared method; we avoid
NeRSemble and AVA-256 because train/test splits differ between methods. Each case
is a camera triplet, and all methods receive the same reference image, driving
video and target calibration; expression is estimated from the driving video
only, and multi-view tracking only places the head under $\pi_{\text{tar}}$.
BecomingLit has 25 self-reenactment cases from 13 identities, with a near-frontal
reference camera and driving and target cameras $35$--$101^\circ$ apart in
azimuth (median $77^\circ$). Its 13 cross-reenactment pairs form one cycle over
the identities; ID is measured against the reference identity, and Expr and
AKD against the driving identity's own video from
$\pi_{\text{tar}}$. Multiface has 22 cases from 11 identities, with driving and
target cameras $40.5$--$90.1^\circ$ apart. VFHQ-test~\cite{xie2022vfhq} adds 44
in-the-wild clips.

\paragraph{3DGS consistency probe.}
The probe has one case per BecomingLit identity (13), frozen at one frame, with
the frontal photo as the only reference and the expression held fixed. The
generative methods follow 81 poses on a closed ellipse around the head (azimuth
$\pm60^\circ$, elevation $+5^\circ$ to $-30^\circ$); ours and MMDM generate the
first 77. For each, we fit a static 3DGS scene~\cite{kerbl20233dgs} (20k
iterations, identical settings) to the generated frames with the exact
trajectory cameras, render it at the 16 real rig cameras, which lie
$0.2$--$11.6^\circ$ (median $2.3^\circ$) from the nearest trajectory pose, and
compare with the captured photos; PSNR and LPIPS are face-region metrics. MVP4D
runs MMVDM from a static camera with the trajectory folded into the head pose, in order to produce smooth renderings.
Its multi-view variant reaches 23.40\,dB PSNR, but flickers significantly, and sits still below ours on all metrics.
UIKA and FlexAvatar are rendered directly; FlexAvatar uses its feed-forward
mode, since its per-subject fitting would have access to the evaluated facial
expression.

\section{Baseline Details}
\label{sec:supp-baselines}

\paragraph{Baselines.}
All methods receive the same reference image, driving video and target
calibration; expression is estimated monocularly from the driving video, and our
multi-view FLAME tracking only places the head under $\pi_{\text{tar}}$.
LAM~\cite{he2025lam} and FlexAvatar~\cite{kirschstein2026flexavatar} are driven
by pixel3DMM~\cite{giebenhain2025pixel3dmm} FLAME2020 fits; FlexAvatar uses its
per-subject inversion (200 steps). UIKA~\cite{wu2026uika} is driven by a
monocular VHAP fit. CAP4D~\cite{taubner2025cap4d} and MVP4D~\cite{mvp4d} use
pixel3DMM fits converted to CAP4D's FLAME basis: \textbf{CAP4D} is the full
pipeline, \textbf{MMDM} samples the target camera directly, and \textbf{MVP4D} is
its video-diffusion model MMVDM at the target camera, since fitting MVP4D's 3DGS
avatar per clip is computationally prohibitive. FaceCam~\cite{facecam}
re-renders a given video under a new camera and takes neither a reference
identity nor an expression signal, so it appears only in the 3DGS probe.

\paragraph{Ablation variants.}
\emph{WAN-A.} runs the frozen WAN-Animate-14B with its own motion encoder and a
ViT-Pose~\cite{vitpose} skeleton detected on the SAM-3D-Body mesh rendered into
$\pi_{\text{tar}}$. 
\emph{WAN-A.\ disent.} passes the same motion codes through
$\net{D}^{\text{WAN}}$ and $\net{E}^{\text{WAN}}$, re-entangled with the mean
camera latent of $\pi_{\text{tar}}$ over the other identities; it uses no X-NeMo
code and no fine-tuning. 

All fine-tuned rows are trained for $5{,}000+250$ steps.
\emph{WAN-A.\ expr.} feeds
WAN-Animate's native code to a frozen $\net{P}_{\text{face}}$; \emph{$+$ frozen
fusers} also freezes $\net{B}_{\text{face}}$. \emph{No disent.~$\dagger$}
replaces $\net{D}^{\text{XN}}$ by a translator from raw X-NeMo to raw
WAN-Animate codes, and \emph{self-driven~$\dagger$} additionally reads the
expression code from the target camera. \emph{No expr.\ codes} zeroes the
expression code, so expression enters only through the FLAME head of the proxy
mesh. \emph{No proxy mesh} removes the proxy mesh and, at test time, uses the
skeleton of the training-free rows. \emph{Rand.\ corpus} draws clips and windows at random
from the same sources.

\section{FactorPortrait Comparison Details}
\label{sec:supp-factorportrait}

FactorPortrait~\cite{tang2025factorportrait} has released neither code nor a
model, so we compare on the self-reenactment videos from its project page,
which contain the reference still, FactorPortrait's output, and the ground
truth, but neither camera parameters nor a separate driving video. Both
methods are scored against the same ground-truth frames with the face-region
metrics of \cref{sec:metrics}, without 3DMM alignment. The setup favors
FactorPortrait: its authors chose the clips, which come from its own capture
system, whereas we have to estimate camera, head, and body by monocular
tracking, and our reference image is extracted from a compressed web video.

\section{Implementation Details}
\label{sec:supp-impl}

We report the configuration of the model evaluated as ``Ours'' in the
BecomingLit self-reenactment experiments and the ablations
(\cref{tab:selfreenact,tab:ablation-self}).

\paragraph{Data and preprocessing.}
The corpus contains 4{,}696 clips of 81 frames from 640 identities:
1{,}846 clips from AVA-256~\cite{martinez2024codecavatarstudio} and 2{,}850
from NeRSemble~\cite{kirschstein2023nersemble}, mostly expression and emotion
sequences. Each clip fixes a camera triplet (\cref{sec:training}); reference and
driving cameras are predominantly near-frontal, whereas target cameras cover a
wide range of viewpoints. The FLAME~\cite{li2017flame} head of the proxy mesh
comes from monocular pixel3DMM~\cite{giebenhain2025pixel3dmm} tracking of the
driving video and is placed per frame by a similarity transform onto the
dataset's multi-view head tracking; torso and neck come from
SAM-3D-Body~\cite{yang2026sam3dbody}, stabilized by pelvis-anchored kinematic
fitting (\cref{sec:supp-seated}). The skeleton projects the keypoints of the
same fitted body into $\pi_{\text{tar}}$ and is drawn with WAN-Animate's own
renderer.

\paragraph{Training and inference.}
We fine-tune 861.5\,M parameters of WAN-Animate-14B~\cite{cheng2025wananimate}
without LoRA (\cref{tab:supp-impl}); the DiT blocks $\net{B}$, the VAE, and the
image and text encoders stay frozen. In our implementation, $\net{T}$ and
$\net{E}^{\text{WAN}}$ with the fixed camera code
$\bar{\mathbf{z}}^{\text{WAN}}_{\text{cam}}$ form a single MLP, and
$\net{P}_{\text{mesh}}$ is initialized to zero, so the proxy mesh has no effect
at initialization. At inference, we integrate $\mathbf{v}_\theta$ with 20 Euler
steps at $576\times1024$.

\begin{table}[t]
\centering
\small
\caption{Fine-tuning configuration. Wall-clock times include periodic
validation.}
\label{tab:supp-impl}
\begin{tabular}{@{}lcc@{}}
\toprule
Trainable module & Parameters & Learning rate \\
\midrule
Face fusers $\net{B}_{\text{face}}$ & 839.0\,M & $5\times10^{-6}$ \\
Face encoder $\net{P}_{\text{face}}$ & 17.8\,M & $10^{-5}$ \\
Translator ($\net{T}$, $\net{E}^{\text{WAN}}$ merged) & 3.1\,M & $10^{-5}$ \\
Expression head of $\net{D}^{\text{XN}}$ & 0.8\,M & $10^{-5}$ \\
$\net{P}_{\text{skel}}$, $\net{P}_{\text{mesh}}$ & 0.3\,M each & $5\times10^{-5}$ \\
\midrule
 & Stage 1 & Stage 2 \\
Resolution & $288\times512$ & $576\times1024$ \\
Steps (wall-clock) & 5{,}000 (25.8\,h) & 250 (8.1\,h) \\
\midrule
\multicolumn{3}{@{}l@{}}{AdamW, weight decay $10^{-3}$, one 81-frame clip per step (77 supervised),}\\
\multicolumn{3}{@{}l@{}}{gradient clipping $1.0$, one A100 (80\,GB).}\\
\bottomrule
\end{tabular}
\end{table}

\section{Proxy-Mesh Conditioning Details}
\label{sec:supp-proxymesh}

\subsection{Pelvis-Anchored Kinematic Fitting}
\label{sec:supp-seated}
SAM-3D-Body~\cite{yang2026sam3dbody} predicts each frame independently.
Aligning these per-frame meshes to the tracked FLAME head with a per-frame
similarity transform fits the face well, but any error in that fit
propagates down the kinematic chain: shoulders and hips swing although the
subject sits still, most visibly from oblique cameras. Temporal smoothing
suppresses the high-frequency jitter but not the low-frequency drift.

Subjects in multi-view face captures are seated, so their pelvis is
stationary. We exploit this by re-expressing the sequence with a
\emph{single} similarity transform $\mathcal{W}$ and per-frame MHR joint
angles $\boldsymbol{\phi}_t$ (6D rotation parameterization), solved jointly
over all $T$ frames:
\begin{equation}
\begin{aligned}
\min_{\mathcal{W},\{\boldsymbol{\phi}_t\}} \sum_{t=1}^{T}
 & \big\lVert \mathcal{W}\big(\mathbf{k}(\boldsymbol{\phi}_t)\big) - \mathbf{l}_t \big\rVert^2
 + \lambda_{\text{pel}} \big\lVert \mathcal{W}\big(\mathbf{j}_{\text{pel}}(\boldsymbol{\phi}_t)\big) - \tilde{\mathbf{j}}_{\text{pel}} \big\rVert^2 \\
 + \,& \lambda_{\text{sh}} \big\lVert \mathcal{W}\big(\mathbf{j}_{\text{sh}}(\boldsymbol{\phi}_t)\big) - \tilde{\mathbf{j}}_{\text{sh}} \big\rVert^2
 + \lambda_{\text{prior}} \lVert \boldsymbol{\phi}_t - \hat{\boldsymbol{\phi}}_t \rVert^2
 + \lambda_{\text{temp}} \lVert \boldsymbol{\phi}_t - \boldsymbol{\phi}_{t-1} \rVert^2 ,
\end{aligned}
\end{equation}
where $\mathbf{k}$ are the MHR face keypoints corresponding to the FLAME
landmarks $\mathbf{l}_t$, $\mathbf{j}_{\text{pel}}$ and
$\mathbf{j}_{\text{sh}}$ the pelvis and shoulder joints with temporal
medians $\tilde{\mathbf{j}}$, and $\hat{\boldsymbol{\phi}}_t$ the SAM-3D
per-frame estimate. The pelvis term encodes the stationary-pelvis assumption; since
the pelvis lies outside the conditioning crop, a light shoulder pin
additionally holds the visible torso in place. Because the MHR forward
kinematics is differentiable, we optimize with Adam for 400 iterations,
using $\lambda_{\text{pel}}{=}1$, $\lambda_{\text{sh}}{=}0.3$,
$\lambda_{\text{prior}}{=}10^{-3}$, $\lambda_{\text{temp}}{=}0.03$.

We validate the fitting on NeRSemble EXP-1 sequences, whose large head
rotations ($37$--$56^\circ$) would expose an over-constrained fit
(\cref{fig:supp-seated}). The pelvis term alone reduces the median pelvis
drift of per-frame alignment from $52$--$79$\,mm to $6$--$12$\,mm; the
shoulder pin then reduces the remaining shoulder wander from $11.2$ to
$0.07$\,mm and frame-to-frame jitter by $14\times$, at an unchanged face
residual ($3.04$ vs.\ $3.06$\,mm). Relative to per-frame alignment, which
is optimal for the face by construction, the face residual rises by at
most $0.8$\,mm: genuine head motion is still tracked, and the constraint
only removes motion the seated subject never performed.

\newsavebox{\seatedfitbox}
\newcommand{\seatedfitrow}[2]{%
  \sbox{\seatedfitbox}{\includegraphics[width=0.93\linewidth]{#2}}%
  \rotatebox{90}{\makebox[\ht\seatedfitbox][c]{\scriptsize #1}}\hspace{2pt}%
  \rotatebox{90}{\makebox[\ht\seatedfitbox][c]{%
    \makebox[0.5\ht\seatedfitbox][c]{\scriptsize Pelvis-anchored (ours)}%
    \makebox[0.5\ht\seatedfitbox][c]{\scriptsize Per-frame}}}\hspace{2pt}%
  \usebox{\seatedfitbox}}
\begin{figure*}[t]
\raggedleft
\seatedfitrow{ID 091, frontal view}{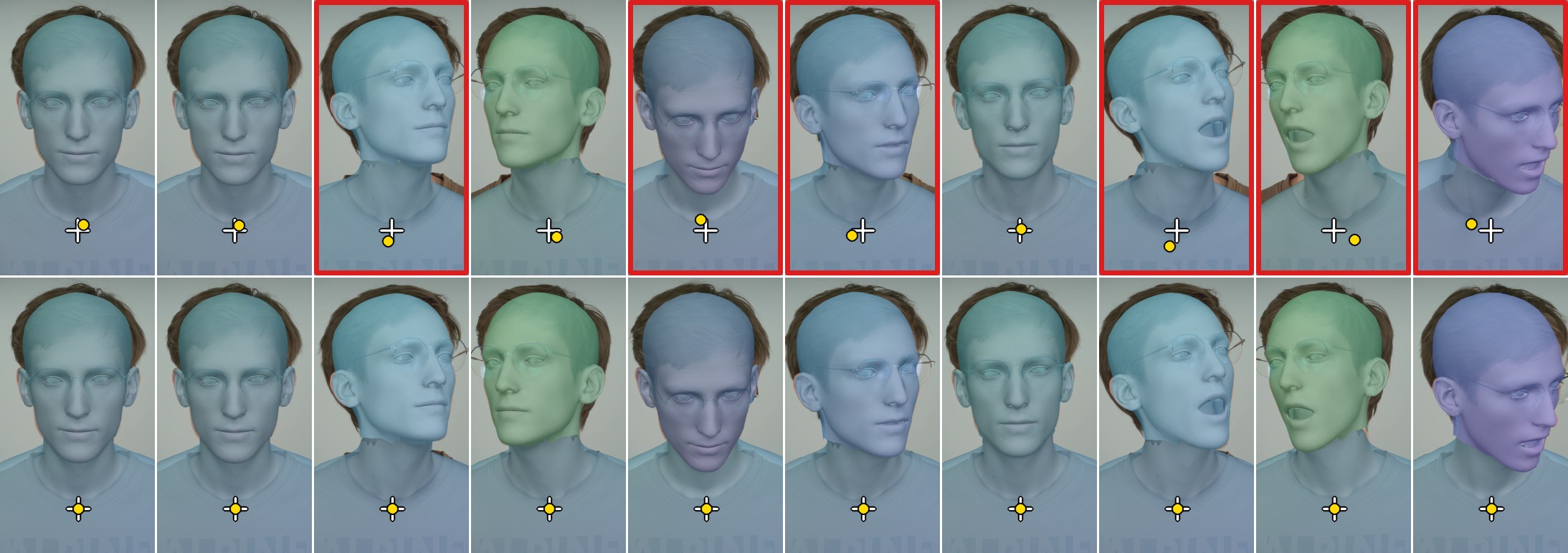}\\[4pt]
\seatedfitrow{ID 285, oblique view}{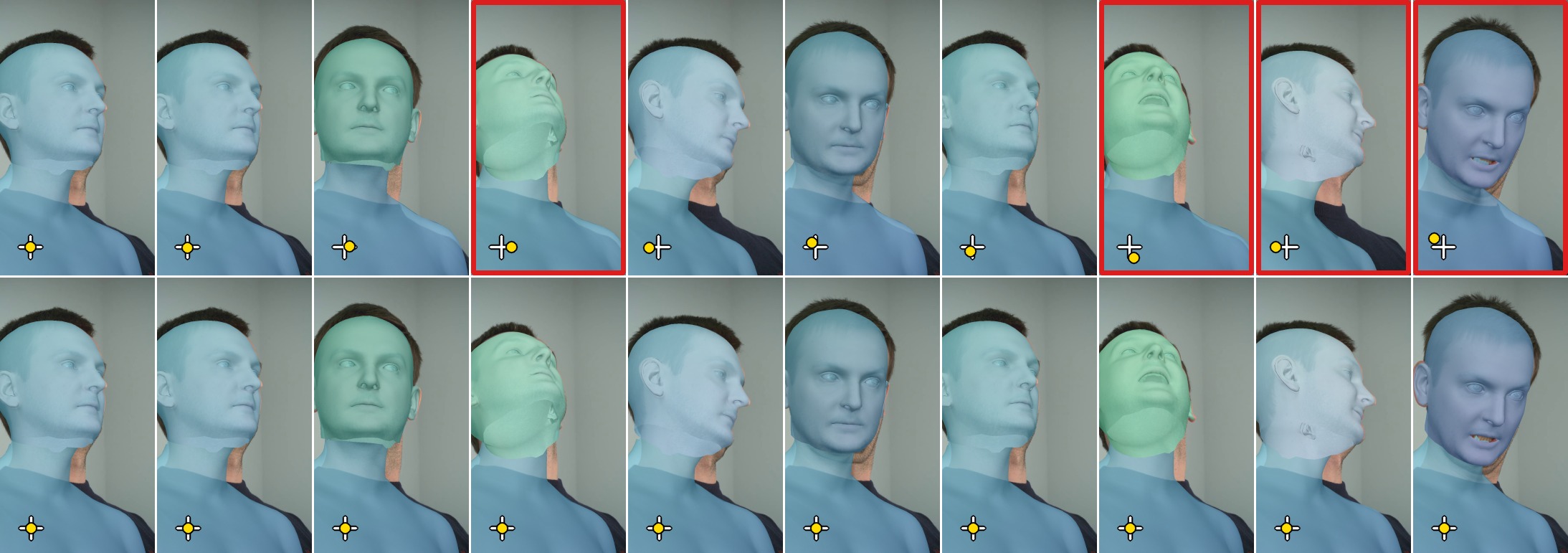}\\[4pt]
\seatedfitrow{ID 901, oblique view}{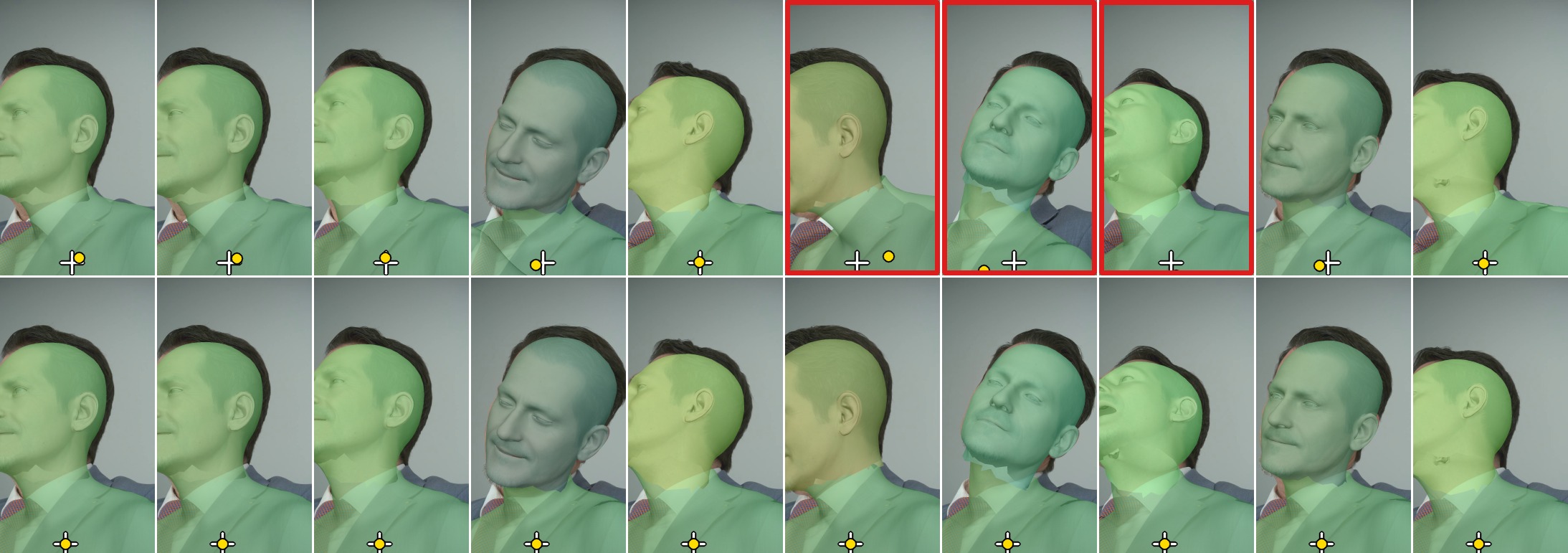}\\[1pt]
\begin{tikzpicture}
  \draw[-{Latex[length=3pt]}, line width=0.4pt] (0,0) -- (0.93\linewidth,0)
    node[midway, below=1pt] {\scriptsize time (10 frames, evenly spaced over the sequence)};
\end{tikzpicture}
\caption{\textbf{Pelvis-anchored kinematic fitting stabilizes the proxy-mesh body.}
Proxy-mesh conditioning renders (\cref{sec:supp-plucker}) overlaid on the
ground-truth frames of three NeRSemble EXP-1 sequences with large head
rotations. For each subject, the top row aligns every per-frame SAM-3D
estimate independently to the tracked FLAME head; the bottom row uses
pelvis-anchored kinematic fitting. Both rows share the same head. The crosshair marks the
shoulder midpoint's median position over the sequence, the yellow dot its
position in the shown frame; red boxes mark frames whose shoulder midpoint
lies more than 45\,px (at $720{\times}1280$) from that median. Under
per-frame alignment the torso drifts,
although the subject is seated; pelvis-anchored fitting holds it in place while the
head, and with it the head's color, follows the performance. The three
camera views show how the ray-direction colors depend on the viewpoint.}
\label{fig:supp-seated}
\end{figure*}

\subsection{Pose-Aware Proxy-Mesh Coloring}
\label{sec:supp-plucker}
For a pixel $\mathbf{u}$ with unit world-space ray direction
$\mathbf{d}(\mathbf{u})$, we color the proxy mesh as
\begin{equation}
  \mathbf{c}(\mathbf{u}) = \tfrac{1}{2}\, g(\mathbf{u})
  + \tfrac{1}{2} \cdot \tfrac{1}{2}\big(\mathbf{R}^{\top}\mathbf{d}(\mathbf{u}) + \mathbf{1}\big),
\end{equation}
where $g$ is Lambertian shading (key light $25^\circ$ above the camera)
and $\mathbf{R}$ is the local frame of the body part covering $\mathbf{u}$.
For the head, $\mathbf{R}_{\text{h}}$ is the rotation that places the
FLAME head in the scene. For the torso, $\mathbf{R}_{\text{b}}$ has its
up axis from the hip midpoint to the neck and its lateral axis between
the hips, orthogonalized. Head and torso are rendered separately and the
head is composited over the torso, so the two regions never interpenetrate.

The choice of frame determines what the color can encode. Camera-space
directions depend only on pixel position and carry no pose information;
world-space directions encode the camera but not the body; and a single
shared frame for both parts leaves a neck rotation visible only through
the silhouette. With per-part local frames, a camera move shifts both
regions' colors together, while a neck rotation changes the head's color
relative to the torso. This ties the neck-control signal directly to the
RGB content of the conditioning. \Cref{fig:supp-seated} illustrates this:
with the torso held in place by pelvis-anchored fitting, its color stays constant
over the sequence, while the head's color changes with every head rotation.

\section{Disentanglement Module Details}
\label{sec:supp-disent}

\section{Disentanglement Training Details}

\subsection{Disentanglement Training}
\label{sec:disentangle_training}

\paragraph{Model.}
Let $\mathbf{z}\in\mathbb{R}^{d}$ be the entangled code of a frozen,
pretrained expression encoder (X-NeMo, $d=512$). The disentangler $D$
consists of independent MLP heads that read the same code in parallel,
\begin{equation}
  D(\mathbf{z})
  = \big(f_{\text{id}}(\mathbf{z}),\, f_{\text{cam}}(\mathbf{z}),\, f_{\text{ex}}(\mathbf{z})\big)
  = (\mathbf{z}_{\text{id}}, \mathbf{z}_{\text{cam}}, \mathbf{z}_{\text{ex}}),
  \label{eq:disentangler}
\end{equation}
with $\mathbf{z}_{\text{ex}}\in\mathbb{R}^{d_{\text{ex}}}$,
$\mathbf{z}_{\text{cam}}\in\mathbb{R}^{d_{\text{cam}}}$ and
$\mathbf{z}_{\text{id}}\in\mathbb{R}^{d_{\text{id}}}$. An optional fourth
head $f_{\text{res}}$ yields a small residual
$\mathbf{z}_{\text{res}}\in\mathbb{R}^{d_{\text{res}}}$ that absorbs what
none of the three supervised factors explains (lighting, crop, background).
The entangler $E$ concatenates the latents and maps them back into the
encoder's space,
\begin{equation}
  E(\mathbf{z}_{\text{id}}, \mathbf{z}_{\text{cam}}, \mathbf{z}_{\text{ex}})
  = g\big([\mathbf{z}_{\text{id}};\,\mathbf{z}_{\text{cam}};\,\mathbf{z}_{\text{ex}};\,\mathbf{z}_{\text{res}}]\big)
  = \hat{\mathbf{z}} .
  \label{eq:entangler}
\end{equation}
All heads $f_{\bullet}$ and $g$ are three-layer ReLU MLPs with a linear
output layer; no latent is normalized, since the norm of $\mathbf{z}$ carries
expression intensity and must survive the round trip.

\paragraph{Supervision from multi-view video.}
We train on NeRSemble~\cite{kirschstein2023nersemble}, where every image is
indexed by identity, sequence, frame, and camera, $(p, s, t, c)$. Codes
$\mathbf{z}$ are pre-computed once for all images of the training identities,
so training operates on cached codes only. Three label functions provide the
supervision:
(i)~the \emph{expression} label $y^{\text{ex}} = (p, s, t)$ -- all 16 cameras
of a frame show the same expression. The human-verified cross-identity pairs
of the DISTRACT identity-invariance task (\cref{sec:benchmark}) are merged
into the same class (transitive closure), so an expression class may span
several identities;
(ii)~the \emph{viewpoint} label $y^{\text{cam}} = c$, the physical camera
(16 classes);
(iii)~the \emph{identity} label $y^{\text{id}} = p$.

\paragraph{Supervised contrastive losses.}
Given a batch $\mathcal{B}=\{\mathbf{z}_i\}_{i=1}^{N}$ with labels $y_i$ and a
head $f$, we apply the supervised contrastive loss~\cite{khosla2020supcon} to
the $\ell_2$-normalized head outputs
$\mathbf{u}_i = f(\mathbf{z}_i)/\lVert f(\mathbf{z}_i)\rVert_2$, with
temperature-scaled similarities $s_{ij} = \mathbf{u}_i^{\!\top}\mathbf{u}_j/\tau$:
\begin{equation}
  \mathcal{L}_{\text{con}}(f, y; \mathcal{B})
  = -\frac{1}{|\mathcal{A}|}\sum_{i\in\mathcal{A}}
    \frac{1}{|P(i)|}\sum_{p\in P(i)}
    \log \frac{\exp(s_{ip})}{\sum_{a\in N(i)} \exp(s_{ia})} ,
  \label{eq:supcon}
\end{equation}
where $N(i) = \{a\neq i\}\setminus\mathcal{I}(i)$ is the set of samples
anchor $i$ is contrasted against, $P(i) = \{p\in N(i) : y_p = y_i\}$ its
positives, $\mathcal{A} = \{i : P(i)\neq\emptyset\}$ the anchors with at least
one positive, and $\mathcal{I}(i)$ an optional
per-anchor ignore set (empty unless stated otherwise). Batches are
class-balanced: we draw $n$ classes and $k$ samples per class. Each head gets
its own loss and its own batch,
\begin{equation}
  \mathcal{L}^{\star}_{\text{con}} = \mathcal{L}_{\text{con}}(f_{\star}, y^{\star}; \mathcal{B}_{\star}),
  \qquad \star \in \{\text{ex}, \text{cam}, \text{id}\}.
  \label{eq:supcon_heads}
\end{equation}
Positives of $f_{\text{ex}}$ are the same expression seen from
\emph{different} cameras (and identities), whereas positives of
$f_{\text{cam}}$ are the same camera across \emph{different} frames and
identities; the three losses thus pull each factor into its own head and push
the other factors out of it.

\paragraph{Reconstruction and residual penalty.}
The entangler must reproduce the original code,
\begin{equation}
  \mathcal{L}_{\text{rec}} = \big\lVert E(D(\mathbf{z})) - \mathbf{z} \big\rVert_1 ,
  \label{eq:recon}
\end{equation}
so the split loses no information; here and below $\lVert\cdot\rVert_1$
denotes the mean absolute error, \ie the $\ell_1$ norm divided by the
dimension. When a residual head is used, an $\ell_1$ penalty
$\mathcal{L}_{\text{res}} = \lVert \mathbf{z}_{\text{res}}\rVert_1$ keeps it
from becoming a shortcut for the reconstruction; its weight
$\lambda_{\text{res}}(e) = \lambda_{\text{res}}\min(1, e/50)$ is ramped in
linearly over the first 50 epochs $e$.

\paragraph{Cycle consistency.}
Reconstruction and contrastive terms alone do not force the entangler to
actually \emph{use} each latent for its designated factor. We therefore swap
latents between two random codes $\mathbf{z}^{a}$ and $\mathbf{z}^{b}$ of the
batch, re-entangle, and require the re-disentangled latents to match:
\begin{align}
  \hat{\mathbf{z}}^{a\leftarrow b} &= E\big(\mathbf{z}^{a}_{\text{id}},\, \mathbf{z}^{b}_{\text{cam}},\, \mathbf{z}^{a}_{\text{ex}}\big), \quad
  \hat{\mathbf{z}}^{b\leftarrow a}  = E\big(\mathbf{z}^{b}_{\text{id}},\, \mathbf{z}^{a}_{\text{cam}},\, \mathbf{z}^{b}_{\text{ex}}\big),
  \label{eq:swap}\\
  \mathcal{L}_{\text{cyc}} &= \tfrac{1}{4}\Big(
      \lVert f_{\text{ex}}(\hat{\mathbf{z}}^{a\leftarrow b}) - \mathbf{z}^{a}_{\text{ex}}\rVert_1
    + \lVert f_{\text{cam}}(\hat{\mathbf{z}}^{a\leftarrow b}) - \mathbf{z}^{b}_{\text{cam}}\rVert_1
    \nonumber\\ &\qquad\;
    + \lVert f_{\text{ex}}(\hat{\mathbf{z}}^{b\leftarrow a}) - \mathbf{z}^{b}_{\text{ex}}\rVert_1
    + \lVert f_{\text{cam}}(\hat{\mathbf{z}}^{b\leftarrow a}) - \mathbf{z}^{a}_{\text{cam}}\rVert_1 \Big)
    \nonumber\\ &\quad
    + \tfrac{1}{2}\Big(
      \lVert f_{\text{id}}(\hat{\mathbf{z}}^{a\leftarrow b}) - \mathbf{z}^{a}_{\text{id}}\rVert_1
    + \lVert f_{\text{id}}(\hat{\mathbf{z}}^{b\leftarrow a}) - \mathbf{z}^{b}_{\text{id}}\rVert_1 \Big).
  \label{eq:cycle}
\end{align}
Gradients flow through both the swapped latents and the re-encoding. The
residual fed to $E$ in \cref{eq:swap} is detached and, per batch, either kept
(each code retains its own $\mathbf{z}_{\text{res}}$), exchanged between $a$
and $b$, or zeroed, with equal probability, so the cycle cannot be closed
through the residual. All cycle terms are switched on at epoch
$e_{\text{cyc}} = 35$, once the heads have stabilized.

expression nor leak into $\mathbf{z}_{\text{ex}}$, and vice versa.

\paragraph{Total objective.}
With $e$ the current epoch, $\gamma_e = \mathbf{1}[e \ge e_{\text{cyc}}]$
gating the cycle terms, and $\star$ running over the three heads, the training
loss is
\begin{equation}
  \begin{split}
  \mathcal{L} ={}& \lambda_{\text{rec}}\,\mathcal{L}_{\text{rec}}
    + \sum_{\star} \lambda_{\star}\,\mathcal{L}^{\star}_{\text{con}}
    + \lambda_{\text{res}}(e)\,\mathcal{L}_{\text{res}} \\
    &+ \gamma_e\Big(
        \lambda_{\text{cyc}}\,\mathcal{L}_{\text{cyc}}
      \Big).
  \end{split}
  \label{eq:total}
\end{equation}

\paragraph{Implementation details.}
For the X-NeMo disentangler we use $d = d_{\text{ex}} = 512$,
$d_{\text{cam}} = 32$, no identity or residual head, and MLP width 512.
Loss weights are $\lambda_{\text{rec}} = \lambda_{\text{ex}} = \lambda_{\text{cam}} = 1$,
$\lambda_{\text{cyc}} = \lambda^{\text{ex}}_{\text{cyc}} = \lambda^{\text{cam}}_{\text{cyc}} = 0.1$
and $\tau = 0.07$ throughout; contrastive batches contain $32 \times 4$
(expression) and $8 \times 16$ (camera) codes. We optimize with AdamW
(learning rate $10^{-3}$, weight decay $10^{-4}$, halved every 200 epochs)
for 1000 epochs with 64 reconstruction codes per step, holding out 10\% of
the identities for validation, and select the checkpoint with the lowest
validation reconstruction error. \cref{tab:disentangle_hparams} lists the
configurations of all disentanglers.

\begin{table}[!htb]
  \centering
\caption{Disentangler configurations. \emph{X-NeMo}: the checkpoint that
  produces $\mathbf{z}_{\text{ex}}$ for all VDA training and inference.
  \emph{WAN-Animate}: the disentangler of WAN-Animate's native expression code
  used for training-free viewpoint control.}
  \label{tab:disentangle_hparams}
  \small
  \setlength{\tabcolsep}{4pt}
  \begin{tabular}{@{}lcc@{}}
    \toprule
     & X-NeMo & WAN-Animate \\
     & \scriptsize(deployed) & \scriptsize(\texttt{epoch\_0850}) \\
    \midrule
    $d$ / $d_{\text{ex}}$ / $d_{\text{cam}}$ & 512 / 512 / 32 & 512 / 512 / 51 \\
    $d_{\text{id}}$ / $d_{\text{res}}$ & -- / -- & 128 / 32 \\
    MLP width (3 layers) & 512 & 512 \\
    $\lambda_{\text{rec}}$ & 1 & 100 \\
    $\lambda_{\text{ex}}$ / $\lambda_{\text{cam}}$ / $\lambda_{\text{id}}$ & 1 / 1 / -- & 1 / 1 / 1 \\
    $\lambda_{\text{res}}$ (ramp 50 ep.) & -- & 10 \\
    $\lambda_{\text{cyc}}$ ($e_{\text{cyc}} = 35$) & 0.1 & 10 \\
    $\lambda^{\text{ex}}_{\text{cyc}}$ / $\lambda^{\text{cam}}_{\text{cyc}}$ & 0.1 / 0.1 & 0.1 / 0.1 \\
    $\tau$ (all heads) & 0.07 & 0.07 \\
    $n\times k$ expression & $32\times4$ & $32\times4$ \\
    $n\times k$ camera & $8\times16$ & $8\times16$ \\
    $n\times k$ identity & -- & $16\times8$ \\
    viewpoint label $y^{\text{cam}}$ & camera & camera \\
    batch (recon.\ codes) & 64 & 64 \\
    lr / weight decay & $10^{-3}$ / $10^{-4}$ & $10^{-3}$ / $10^{-4}$ \\
    lr schedule & step $\times0.5$ / 200 & step $\times0.5$ / 200 \\
    epochs (selected) & 1000 (700) & 1000 (850) \\
    \bottomrule
  \end{tabular}
\end{table}

\section{Training-Free Use of the Disentangled Codes}
\label{sec:supp-disent-apps}

The disentangler of \cref{sec:benchmark} splits an expression encoder's
raw code into an expression latent $\mathbf{z}_{\text{ex}}$ and a camera
latent $\mathbf{z}_{\text{cam}}$, and the entangler maps any pair back into
the encoder's native code space. Both generators below stay frozen: we only
change the code they receive.

\begin{figure*}[!htb]
\centering
\includegraphics[width=0.7\linewidth]{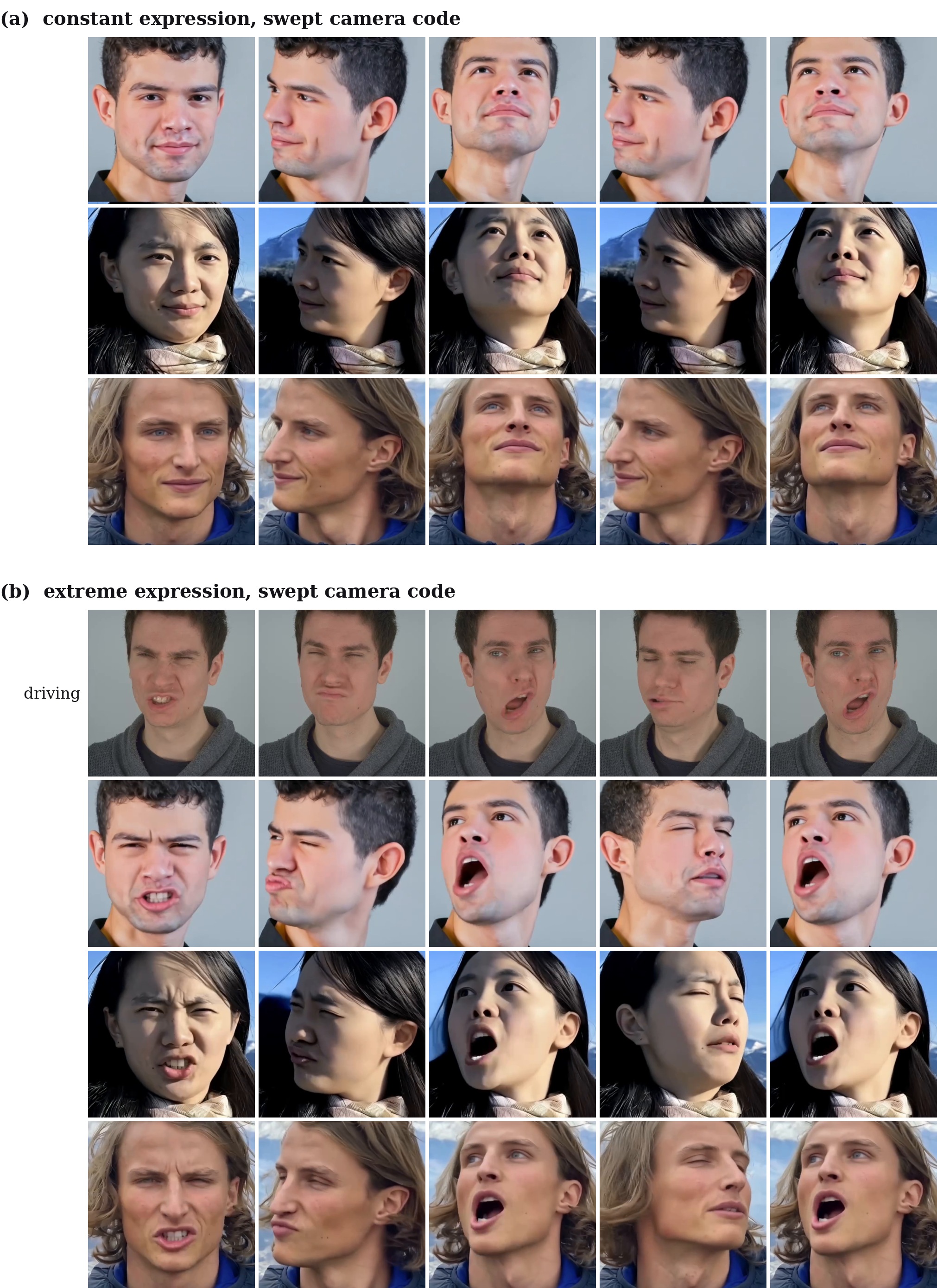}
\caption{Training-free head-pose control in X-NeMo. The disentangled
expression code is kept fixed while the camera code is swept through other
views, (a) with a constant expression and (b) with an extreme driving
expression (top row: driving frames). The head rotates and the expression is
preserved; X-NeMo does not separate camera from neck motion, so the camera
code moves the head rather than the viewpoint.}
\label{fig:xnemo-headpose}
\end{figure*}

\paragraph{X-NeMo.} Keeping $\mathbf{z}_{\text{ex}}$ fixed and sweeping
$\mathbf{z}_{\text{cam}}$ through the codes of other camera views rotates the
generated head while the expression is preserved, both for a constant and for
an extreme driving expression (\cref{fig:xnemo-headpose}). X-NeMo cannot
distinguish a camera change from a neck rotation, so the camera code moves the
head rather than the viewpoint. This is why our avatar model takes camera and
body pose from the proxy mesh instead (\cref{sec:poseconditioning}).

\begin{figure*}[!htb]
\centering
\includegraphics[width=\linewidth]{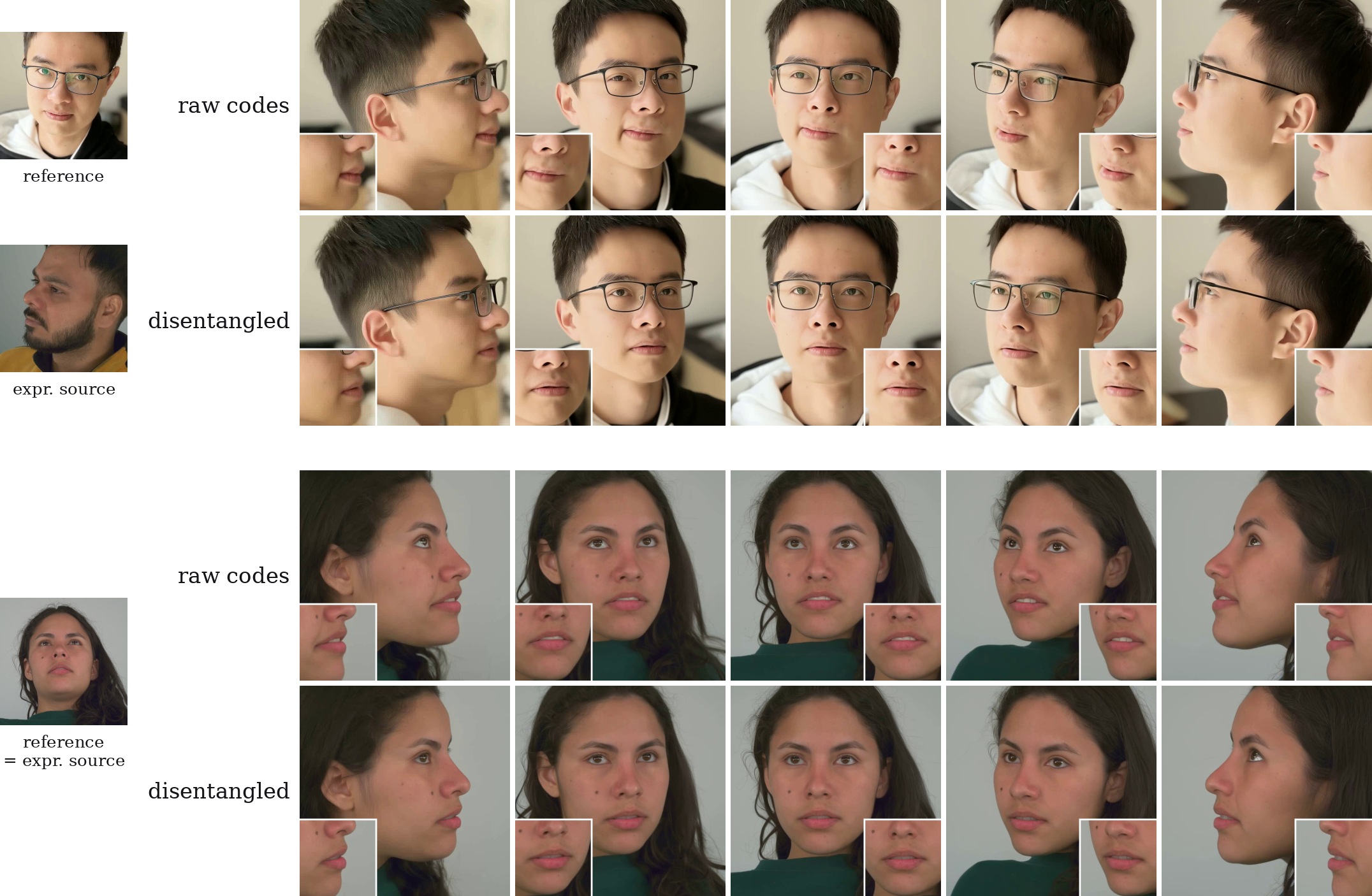}
\caption{Frozen WAN-Animate rendering five views of a camera orbit, driven by
its raw expression codes or by our disentangled codes (expression latent
re-entangled with the target view's camera code). Insets magnify the mouth.
Top: the expression source is a profile view of another person. Bottom: the
expression is taken from the reference image itself. Raw codes carry the
source viewpoint, which shows up as expression artifacts once the target view
differs; the disentangled codes preserve the source expression.}
\label{fig:wan-disent-nvs}
\end{figure*}

\paragraph{WAN-Animate.} WAN-Animate's native expression codes carry the
viewpoint of the frame they are extracted from. When the target view differs
from the expression source, rendering with the raw codes produces subtle but
visible expression artifacts (parted lips, a shifted mouth shape) that vary
along the camera path. Re-entangling our disentangled expression latent with
the target view's camera code removes them (\cref{fig:wan-disent-nvs}). The
remaining artifacts of this training-free variant, and the coarse camera
control it offers, motivate the camera-triplet fine-tuning of
\cref{sec:method}.

\csname iclr@appsection\endcsname{Emergent View Invariance}
\label{sec:supp-emergence}

\begin{figure}[!htb]
\centering
\includegraphics[width=\columnwidth]{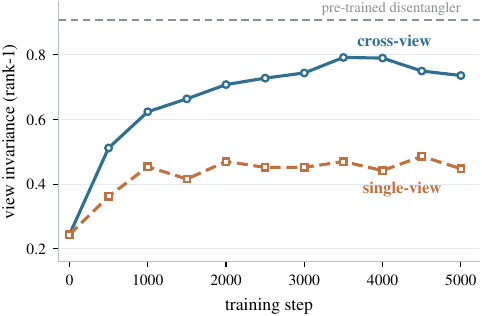}
\caption{\textbf{Cross-view supervision creates view invariance; single-view supervision does not.} DISTRACT viewpoint-invariance rank-1 of the expression code at the face-encoder output, over 500 triplets, for two runs that differ only in the camera the driving expression code is read from. Both begin at 0.244, WAN-Animate's pre-trained face encoder on the shared translator initialisation. Cross-view training reaches 0.79, while single-view training stalls below 0.49. The final gap is 0.288.}
\label{fig:disent-emergence}
\end{figure}

Camera-triplet supervision alone makes the expression code view-invariant.
We compare the two runs marked $\dagger$ in \cref{tab:ablation-self}, which
differ only in whether the driving expression code is read from a different
camera than the reconstructed frame (cross-view) or from the same camera
(self-driven), and track DISTRACT viewpoint-invariance rank-1 accuracy of the
expression code at the face-encoder output during training
(\cref{fig:disent-emergence}). Both start at 0.244; cross-view training climbs
to 0.792, whereas self-driven training stalls at 0.486. Since the proxy mesh
already fixes the target camera, viewpoint information in the expression code
is never needed; under cross-view supervision it is actively misleading, and
training suppresses it. The gain carries over to generation quality
(\cref{tab:ablation-self}).

\section{Streaming Distillation Prototype}
\label{sec:supp-streaming}

As a proof of concept for fast and long generation, we distill our model with
Self Forcing~\cite{huang2025selfforcing} into a causal, four-step student.
The student shares our architecture, weights, and conditioning; we only train
rank-128 LoRA adapters on the transformer and the face-fuser blocks, and make
attention block-causal over blocks of four latent frames (16 video frames).
Training follows the two stages of Self Forcing, an ODE
initialization~\cite{yin2025causvid} on 1{,}500 teacher trajectories followed
by distribution matching distillation~\cite{yin2024dmd,yin2024dmd2} on
self-generated videos, at $288\times512$ and takes about 60 A100 GPU hours.
At inference, the student streams the video block by block with a rolling
key--value cache and two frame sinks~\cite{xiao2024streamingllm,yang2026longlive},
the reference latent and the first generated frames, so that memory is
constant in the video length.

\begin{figure*}[!htb]
\centering
\includegraphics[width=\linewidth]{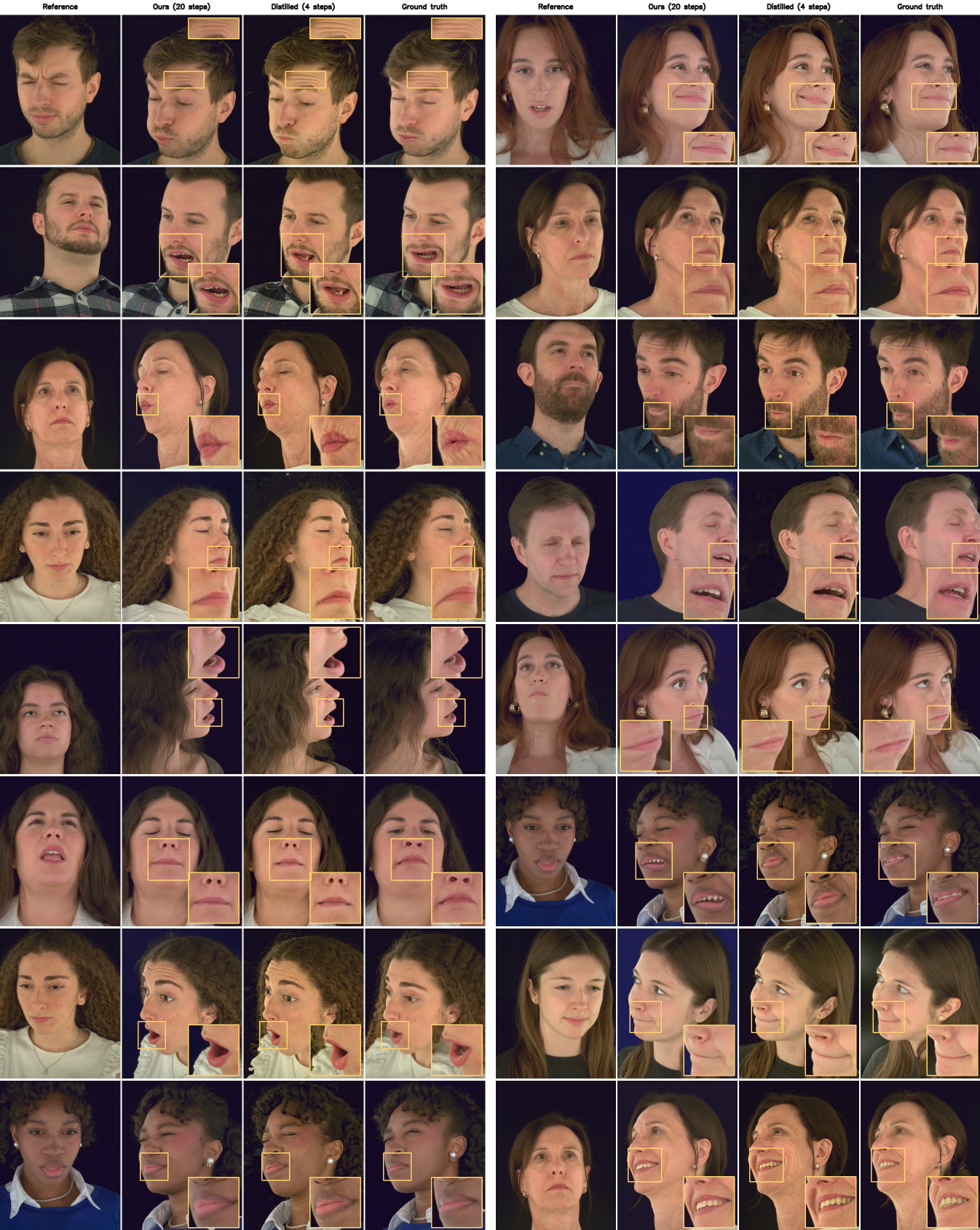}
\caption{Our full model (20 steps) versus the distilled four-step student on
BecomingLit self-reenactment. Insets magnify facial details.}
\label{fig:supp-distill}
\end{figure*}

On one A100, the student generates 4.8 frames per second at $288\times512$: a
41-second video (1029 frames) takes about four minutes.
\Cref{tab:supp-distill,fig:supp-distill} compare the student with our full model on
BecomingLit self-reenactment: it denoises a clip about $11\times$ faster and
matches identity, expression, and keypoint accuracy at a moderate loss in
image quality. On
41-second videos of 11 held-out subjects, identity and appearance remain
stable to the last frame, and expression fidelity stays constant while chunked
generation in overlapping windows degrades over time. The student is somewhat
behind our full model in image quality and runs below real time; training at higher resolution and system-level optimizations are
natural next steps.

\begin{table}[!htb]
\centering
\caption{Full model versus the distilled four-step student on BecomingLit
self-reenactment ($n$=25, $576\times1024$). -F: face region, as in
\cref{tab:selfreenact}. Denoising time per clip excludes conditioning
preprocessing and VAE decoding.}
\label{tab:supp-distill}
\small
\setlength{\tabcolsep}{3pt}
\begin{adjustbox}{max width=\linewidth}
\begin{tabular}{@{}lccccccccc@{}}
\toprule
Method & Steps & \begin{tabular}[b]{@{}c@{}}Network calls\\per clip\end{tabular} & \begin{tabular}[b]{@{}c@{}}Denoising time\\per clip\end{tabular} & PSNR-F~$\uparrow$ & LPIPS-F~$\downarrow$ & JOD-F~$\uparrow$ & ID~$\uparrow$ & Expr~$\uparrow$ & AKD~$\downarrow$ \\
\midrule
Ours (full model)   & 20 & 40 & 21.2\,min & \textbf{22.43} & \textbf{0.202} & \textbf{3.99} & 0.699 & \textbf{0.920} & 0.055 \\
Distilled (4 steps) & 4  & 4  & \textbf{1.9\,min} & 21.81 & 0.217 & 3.57 & \textbf{0.714} & 0.917 & \textbf{0.051} \\
\bottomrule
\end{tabular}
\end{adjustbox}
\end{table}

\csname iclr@appsection\endcsname{Full Quantitative Results}
\label{sec:supp-tables}

\Cref{tab:distract-full} reports all evaluated encoders on DISTRACT without the
MLP probe.

\begin{table}[!htb]
\centering
\caption{All 8 evaluated encoders on the DISTRACT benchmark ($n$=500 for Task~1, $n$=126 for Task~2; entries in \%), without the MLP probe. Columns as in \cref{tab:distract}. Classical 3DMM regressors (top) are consistently less viewpoint-biased than the learned reenactment/portrait-animation encoders below the rule.}
\label{tab:distract-full}
\scriptsize
\setlength{\tabcolsep}{0.6pt}
\setlength{\fboxsep}{0.6pt}
\begin{tabular}{lcccc}
\toprule
 & \multicolumn{2}{c}{View Inv.} & \multicolumn{2}{c}{Identity Inv.} \\
\cmidrule(lr){2-3} \cmidrule(lr){4-5}
Encoder & R-1$\uparrow$ & Err A/B/C/D$\downarrow$ & R-1$\uparrow$ & Err A/B/C/D$\downarrow$ \\
\midrule
{\tiny DECA~\cite{feng2021deca}}                       & \cellcolor{tabfirst}{61} & \colorbox{tabfirst}{33}/\colorbox{tabsecond}{18}/\colorbox{tabsecond}{18}/\colorbox{tabsecond}{11} & \cellcolor{tabfirst}{34} & \colorbox{tabfirst}{51}/\colorbox{tabthird}{37}/\colorbox{tabfirst}{25}/\colorbox{tabthird}{24} \\
{\tiny EMOCA~\cite{danecek2022emoca}}                  & \cellcolor{tabthird}{42} & \colorbox{tabthird}{47}/\colorbox{tabthird}{22}/\colorbox{tabthird}{30}/\colorbox{tabthird}{16} & \cellcolor{tabsecond}{32} & \colorbox{tabsecond}{56}/\colorbox{tabsecond}{34}/\colorbox{tabthird}{43}/\colorbox{tabfirst}{21} \\
{\tiny SHeaP~\cite{schoneveld2025sheap}}               & \cellcolor{tabsecond}{54} & \colorbox{tabsecond}{41}/\colorbox{tabfirst}{17}/\colorbox{tabfirst}{07}/\colorbox{tabfirst}{03} & \cellcolor{tabthird}{17} & \colorbox{tabthird}{69}/{52}/\colorbox{tabsecond}{31}/\colorbox{tabsecond}{21} \\
\midrule
{\tiny EMOPortraits~\cite{drobyshev2024emoportraits}}  & {23} & {70}/{37}/{51}/{28} & {14} & {76}/{42}/{58}/{36} \\
{\tiny PD-FGC~\cite{wang2023pdfgc}}                    & {08} & {90}/{42}/{79}/{43} & {07} & {91}/{48}/{81}/{43} \\
{\tiny VOODOO-XP~\cite{tran2024voodooxp}}              & {15} & {84}/{36}/{55}/{28} & {06} & {94}/{56}/{71}/{39} \\
{\tiny WAN-Animate~\cite{cheng2025wananimate}}         & {21} & {72}/{34}/{59}/{33} & {14} & {74}/\colorbox{tabfirst}{32}/{62}/{26} \\
{\tiny X-NeMo~\cite{zhao2025xnemo}}                    & {18} & {78}/{35}/{70}/{34} & {14} & {78}/\colorbox{tabthird}{37}/{68}/{37} \\
\bottomrule
\end{tabular}
\end{table}

\section{Additional Qualitative Results}
\label{sec:supp-qualitative}\label{sec:supp-figures}

\begin{figure*}[t]
\centering
\includegraphics[width=\linewidth]{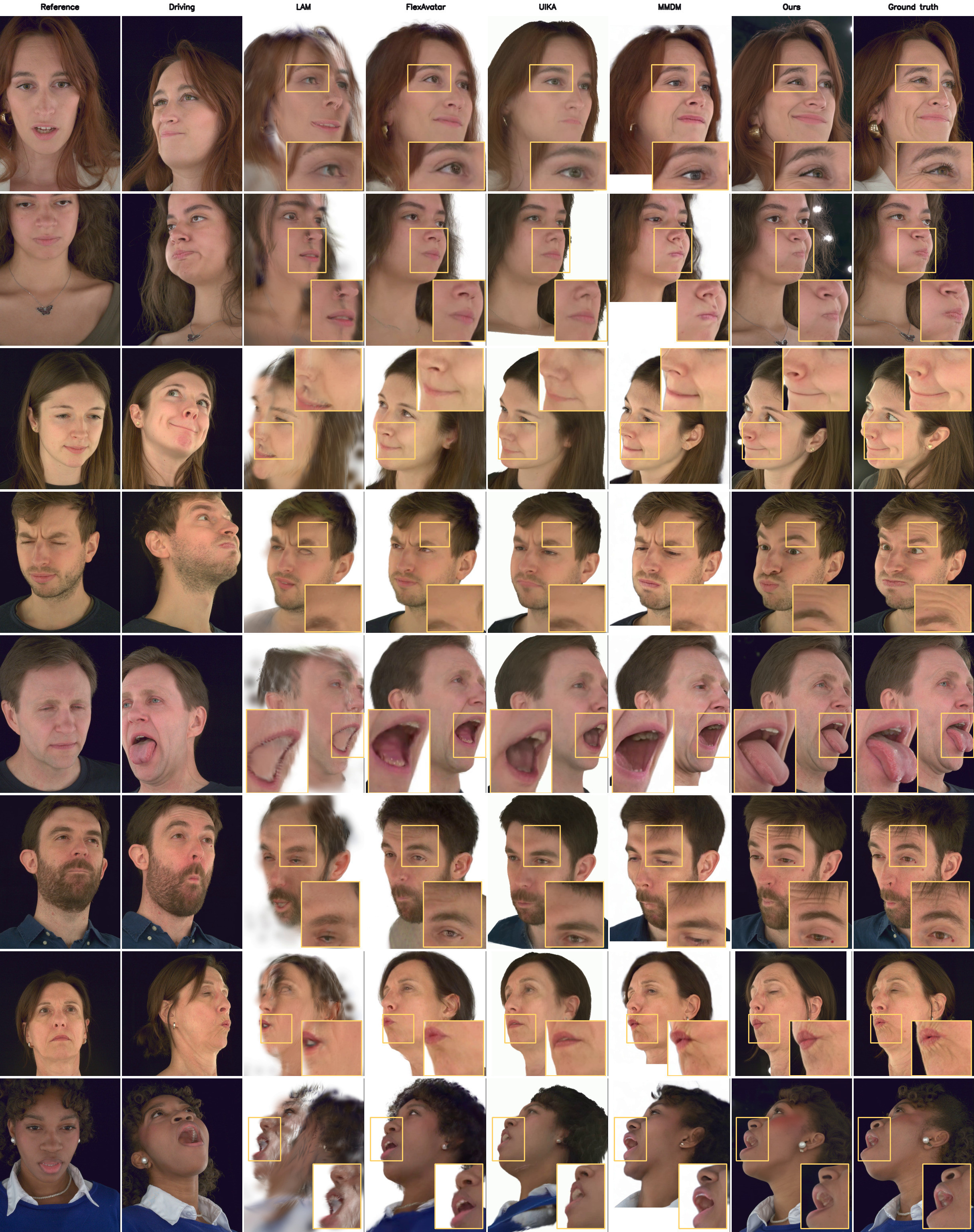}
\caption{BecomingLit self-reenactment, qualitative comparison
of our method against LAM~\cite{he2025lam},
FlexAvatar~\cite{kirschstein2026flexavatar}, UIKA~\cite{wu2026uika}, and MMDM
(the morphable multi-view diffusion component of
CAP4D~\cite{taubner2025cap4d}, rendered directly at the target camera -- not
the full CAP4D avatar pipeline). One row per identity (8 of the benchmark's
13), each frame chosen to maximize expression diversity within that
identity's own driving clips (farthest-point sampling over X-NeMo expression
embeddings~\cite{zhao2025xnemo}), then narrowed from an initial per-identity
diverse set down to these 8 through several rounds of manual review.}
\label{fig:selfreenact-becominglit}
\end{figure*}

\begin{figure*}[p]
\centering
\includegraphics[width=0.82\linewidth]{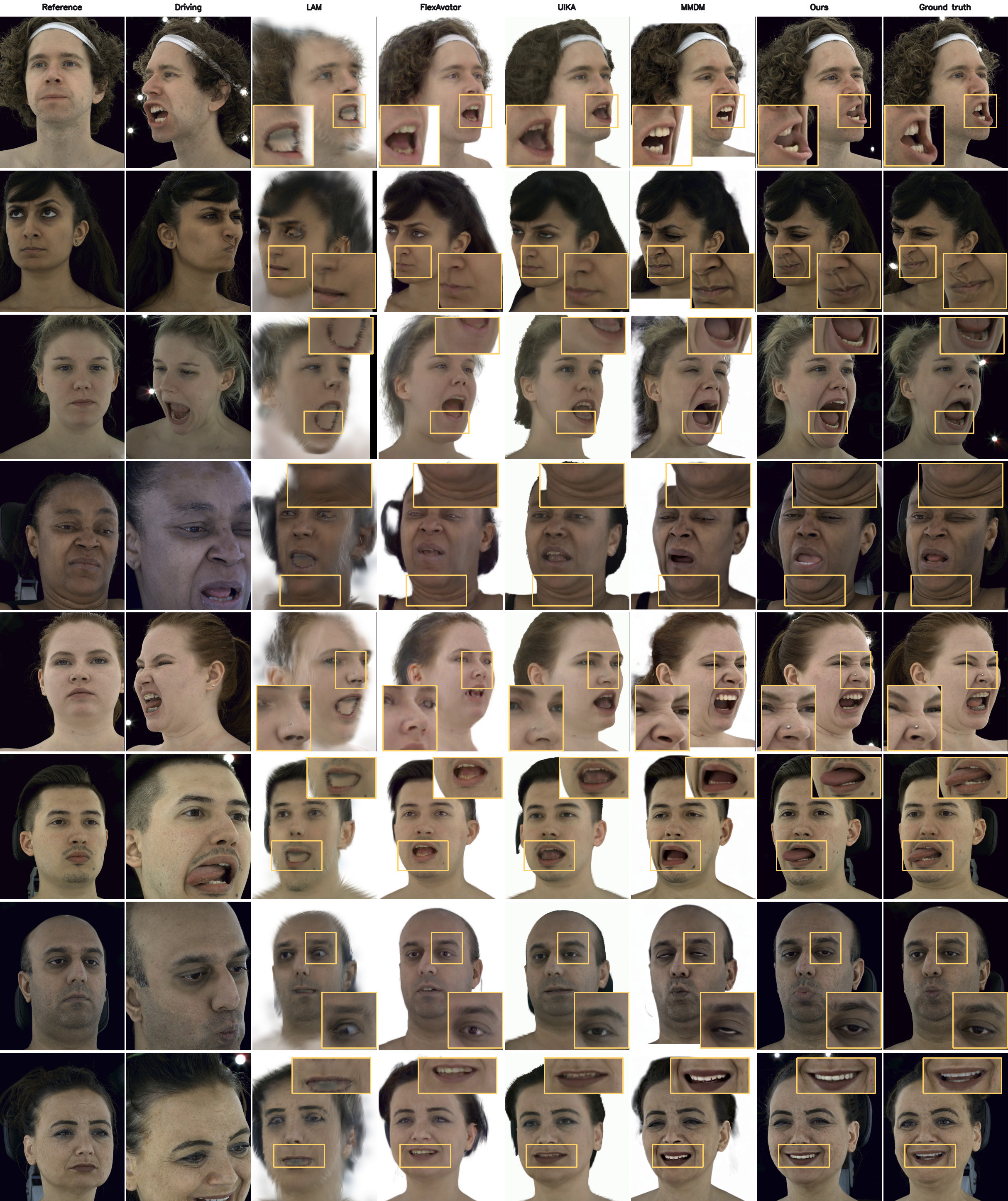}
\caption{Multiface self-reenactment, qualitative comparison
of our method against LAM~\cite{he2025lam},
FlexAvatar~\cite{kirschstein2026flexavatar},
UIKA~\cite{wu2026uika}, and MMDM
(the morphable multi-view diffusion component of
CAP4D~\cite{taubner2025cap4d}, rendered directly at the target camera -- not
the full CAP4D avatar pipeline).
One row per identity (8 of the benchmark's 11), each frame chosen to
maximize expression diversity within that identity's own driving clips
(farthest-point sampling over X-NeMo expression
embeddings~\cite{zhao2025xnemo}), then narrowed from an initial per-identity
diverse set down to these 8 through two rounds of manual review.}
\label{fig:selfreenact-multiface}
\end{figure*}

We encourage the reader to watch our \vdasuppvideo. 
Full 8-row self-reenactment comparisons on BecomingLit
(\cref{fig:selfreenact-becominglit}) and Multiface (\cref{fig:selfreenact-multiface});
\cref{fig:selfreenact} in the main paper shows a subset of these rows.
\Cref{fig:selfreenact-vfhq} shows qualitative comparisons on VFHQ-test.

\begin{figure}[t]
\centering
\includegraphics[width=\linewidth]{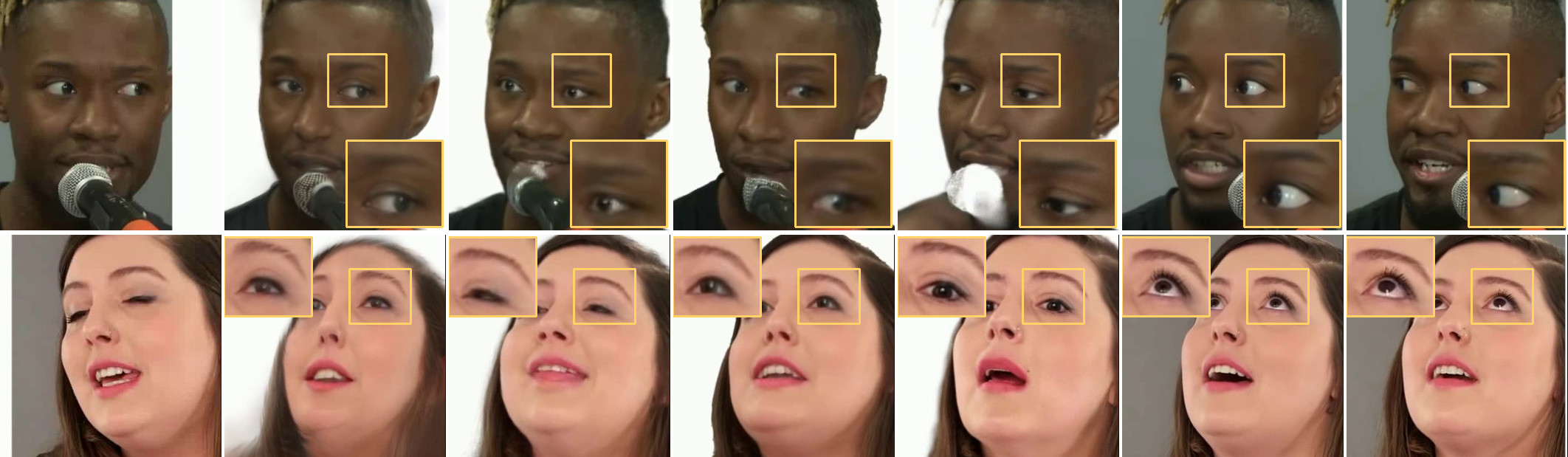}\\[2pt]
{\tiny
\makebox[0.1428\linewidth]{Ref.}%
\makebox[0.1428\linewidth]{LAM}%
\makebox[0.1428\linewidth]{FlexAvatar}%
\makebox[0.1428\linewidth]{UIKA}%
\makebox[0.1428\linewidth]{CAP4D}%
\makebox[0.1428\linewidth]{Ours}%
\makebox[0.1428\linewidth]{GT}}
\caption{Qualitative comparisons on the in-the-wild VFHQ test
set~\cite{xie2022vfhq}, against SotA 3D avatars.}
\label{fig:selfreenact-vfhq}
\end{figure}

\end{document}